\documentclass{article}

\usepackage[final,main]{neurips_2026}

\usepackage[utf8]{inputenc}
\usepackage[T1]{fontenc}
\usepackage[table]{xcolor}
\usepackage{hyperref}
\usepackage{url}
\usepackage{graphicx}
\usepackage{subcaption}
\usepackage{booktabs}
\usepackage{makecell}
\usepackage{array}
\usepackage{amsmath}
\usepackage{amssymb}
\usepackage{amsfonts}
\usepackage{mathtools}
\usepackage{amsthm}
\usepackage{nicefrac}
\usepackage{float}
\usepackage{afterpage}
\usepackage{microtype}
\usepackage{placeins}
\usepackage[capitalize,noabbrev]{cleveref}

\let\cite\citep

\renewcommand{\arraystretch}{1.12}
\newcommand{\twoline}[2]{\makecell[c]{#1\\[-1pt]\scriptsize #2}}

\newcolumntype{C}[1]{>{\centering\arraybackslash}p{#1}}

\theoremstyle{plain}

\theoremstyle{definition}

\theoremstyle{remark}

\title{Beyond Selection: Token Parameterization for Extreme Visual Token Compression}

\hypersetup{
  hidelinks,
  pdftitle={Beyond Selection: Token Parameterization for Extreme Visual Token Compression},
  pdfauthor={Rui Zhong, Yu Li, Zheyu Yan, Cheng Zhuo}
}

\author{%
  Rui Zhong \quad Yu Li \quad Zheyu Yan\thanks{Corresponding author: \texttt{zyan2@zju.edu.cn}.\quad \protect\hypertarget{code-release}{\textsuperscript{\(\dagger\)}}Code: \url{https://github.com/zrrraa/Braco}.} \quad Cheng Zhuo \\
  Zhejiang University \\
  \texttt{\{rzhong,li.yu,zyan2,czhuo\}@zju.edu.cn} \\
}

\begin{document}

\maketitle

\begin{abstract}
Visual-token compression is effective for improving the efficiency of vision-language models, but under extreme compression budgets, token pruning can break visual grounding while learned resamplers increase parameter count, attention cost, and training complexity.
We revisit compression through a \textbf{token parameterization} lens, separating (i) \emph{basis transformation and
structured truncation} (retained subspace/compressibility) from (ii) \emph{coordinate organization} (optimization and
cross-modal alignment).
This view yields two coupled objectives, \emph{compressibility} and \emph{learnability}, which we formalize as unified
functionals.
Guided by these objectives, we design Braco\hyperlink{code-release}{\textsuperscript{\(\dagger\)}}, a lightweight four-step coder that combines transform-basis truncation,
input-independent basis-coordinate embeddings, budget-dependent orthogonal re-parameterization, and learned spatial residual
tokens from lightweight pooling.
Experiments show that Braco forms the favorable empirical accuracy-efficiency frontier under $23\times$--$64\times$
compression and remains competitive at $144\times$, reaching $95.2\%$ accuracy while reducing prefill FLOPs by
84.2\%--86.7\% relative to the uncompressed upper bound.
Against prior methods, Braco matches or improves accuracy while achieving up to a $\sim$36\% end-to-end speedup and using
16.6$\times$/78.8$\times$ lower compressor latency/FLOPs.
\end{abstract}

\section{Introduction}
\label{sec:intro}

\afterpage{%
\begin{figure}[t]
    \centering
    \includegraphics[width=\linewidth]{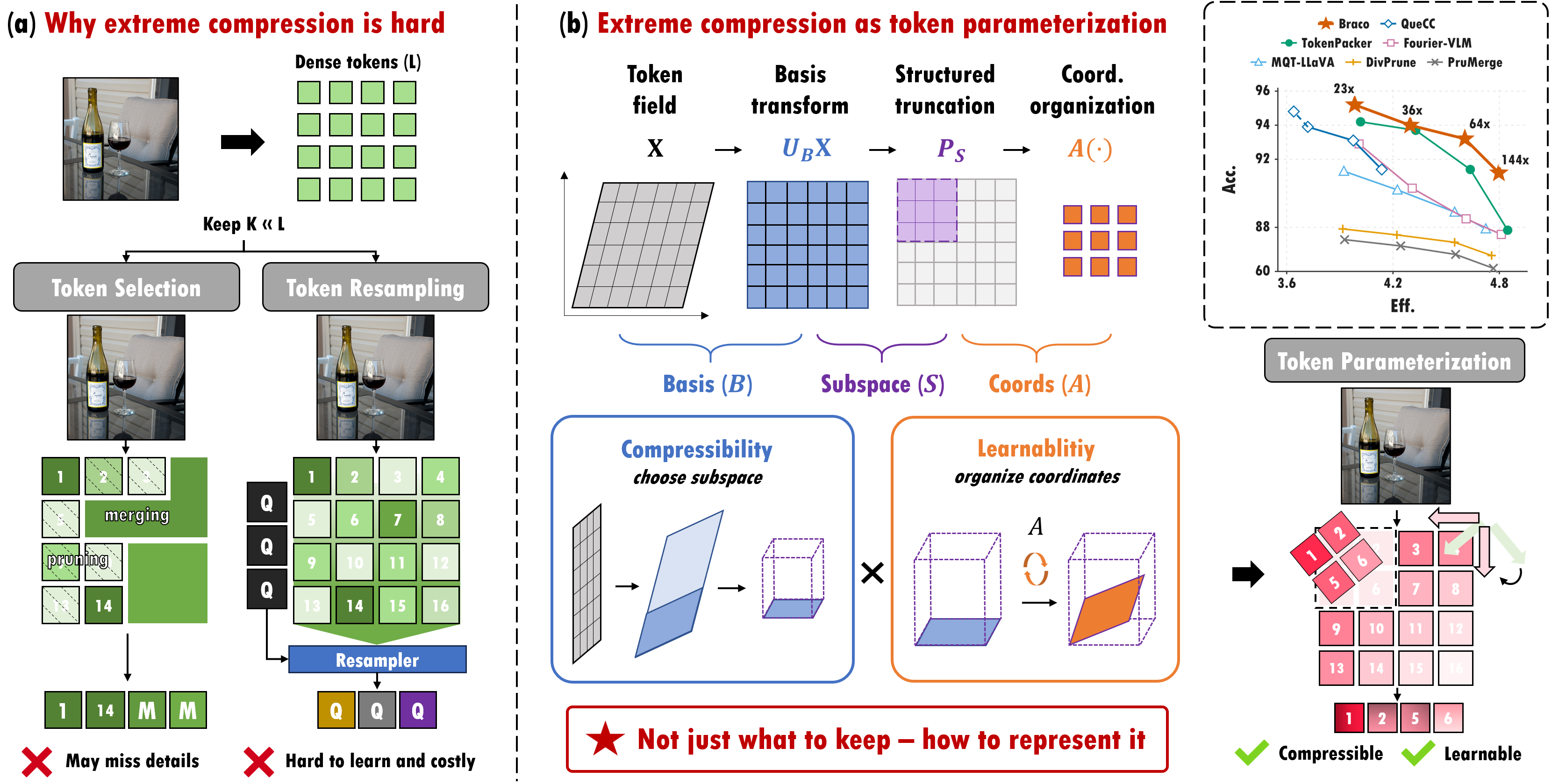}
    \vspace{-2mm}
\caption{\textbf{Overview.}
Token selection can miss rare evidence and learned resampling can be costly; Braco instead parameterizes the visual token
field by choosing a basis/subspace for compressibility, organizing coordinates for learnability, and adding a lightweight
residual branch.
The right plot summarizes the empirical accuracy--efficiency frontier. Acc. is benchmark accuracy normalized by the uncompressed
interface, and Eff. is the average of FLOPs- and latency-efficiency ratios.}
\vspace{-3mm}
    \label{fig:overview}
\end{figure}
}

Vision--language models (VLMs) encode images as visual-token sequences, enabling strong multimodal
reasoning~\cite{bai2025qwen3vltechnicalreport,vteam2026glm45vglm41vthinkingversatilemultimodal,wu2024deepseek,zhu2025internvl3}
but incurring costs linear in token count.
Although many systems use hundreds or thousands of image tokens, deployment settings such as mobile inference,
low-latency interactive agents, and long-context multimodal reasoning may allow only dozens.
In this regime, naive token reduction often degrades grounding and compositional understanding, creating a sharp
efficiency--fidelity trade-off.

Prior work therefore compresses visual tokens before or within the LLM.
The most straightforward method is \emph{token selection}: pruning unimportant patches or merging redundant ones~\cite{rao2021dynamicvit,liang2022not,bolya2022token,alvar2025divprune}.
These methods offer strong performance--latency trade-offs at moderate compression with low overhead.
Under \emph{extreme} compression, they become brittle by dropping rare but critical regions.
Unstable importance proxies such as raw attention~\cite{jain2019attention} further weaken this strategy at high
compression ratios~\cite{alvar2025divprune,wen2025stop,zhang_beyond_2025-1}.
Learned resampling mitigates this brittleness by distilling dense grids into fixed latent tokens through attention
bottlenecks such as Perceiver-style cross-attention and query-driven
interfaces~\cite{jaegle2021perceiver,alayrac2022flamingo,li2023blip,zhang2025llava,li2025tokenpacker,li2024inference}.
This accuracy often comes with extra attention computation or parameters inside the compression module, as well as staged
training and careful multimodal alignment~\cite{alayrac2022flamingo,li2023blip}; \cref{fig:overview} contrasts these
failure modes.

Taken together, performance at extreme compression ratios is limited by two constraints.
The retained subspace must concentrate task-relevant information (\emph{compressibility}), and the retained coordinates must
induce a tractable alignment problem (\emph{learnability}).
These observations motivate a basic question: under extreme token budgets, is performance mainly a matter of selecting
tokens more carefully, or of choosing a better \emph{parameterization} for the visual token field?

We address this question by treating extreme vision-token compression as a \textbf{token parameterization} problem.
Under this view, we build \textit{Braco} (\textbf{B}ackbone--\textbf{r}esidual + b\textbf{a}sis + \textbf{co}ordinate),
a deployable coder with four components.
\textbf{(1) Basis choice:} we \emph{re-express} the spatial token field in an orthonormal transform basis to concentrate
task-relevant information into a compact coefficient set, then apply structured truncation under a fixed budget.
\textbf{(2) Position embedding:} because transform coefficients lose explicit spatial identity, we inject an
input-independent \emph{basis-coordinate} embedding on the transform lattice to provide consistent indexing cues for
downstream multimodal fusion.
\textbf{(3) Coordinate organization:} within the \emph{same} retained subspace, we optionally apply an orthogonal
re-parameterization that trades statistical conditioning against geometric compatibility, since downstream optimization is
not invariant to token-axis rotations~\cite{salimans2016weight}.
\textbf{(4) Residual connection:} finally, we add a small set of learned spatial residual tokens via lightweight sparse
pooling to recover localized details beyond the low-pass backbone.
This yields a factorized compressed interface: a compact transform-domain backbone, coordinate identity for alignment, and
sparse spatial residuals for local cues beyond the low-pass subspace.

In summary, we make the following contributions:

\begin{itemize}
    \item We introduce a token-parameterization view of extreme visual-token compression. 
    Rather than treating compression as token selection alone, we separate basis transformation and structured truncation, which determine the retained subspace, from coordinate organization, which affects optimization and cross-modal alignment.
    \item We formalize this view through unified compressibility and learnability diagnostics. 
    These objectives quantify how well a structured subspace preserves task-relevant information and how coordinate choices within the same subspace affect conditioning and downstream alignment.
    \item We compensate for transform truncation with basis-coordinate embeddings
    and spatial residuals.
    We restore stable token identity in the transform lattice using an input-independent
    basis-coordinate embedding, and we compensate for information lost under low-pass
    truncation using a lightweight sparse pooling module that learns a small set of spatial
    residual tokens.
    \item We design a lightweight, deployable four-step coder.
    Guided by the analysis, we design Braco, which combines transform-basis truncation,
    basis-coordinate embeddings, subspace-preserving re-parameterization, and a small
    spatial residual budget to recover localized details.
\end{itemize}

Experiments show that Braco realizes a strong empirical accuracy--efficiency frontier under extreme budgets (\cref{fig:overview}).  At
matched budgets, it gives the leading aggregate Acc.--cost trade-off at 25/16/9 tokens; under $36\times$--$64\times$
compression, it matches a competitive prior while running $\sim$36\% faster.  At 16 tokens, its compressor has 16.6$\times$ lower latency
and 78.8$\times$ fewer FLOPs than that prior compressor.  Relative to the 576-token Vanilla upper bound, Braco
preserves $95.2\%$ Acc.\ at $23\times$ and $93.2$--$94.0\%$ Acc.\ at $36\times$--$64\times$ while reducing prefill FLOPs
to 1.15--1.37T; on larger inputs, it retains 90.8\% Acc.\ at $182\times$ versus 68.9\% for a comparable prior.  These
results show that extreme compression needs a compact, alignable visual coordinate system, not just fewer tokens.

\section{Related Work}
\label{sec:related}

\noindent\textbf{Token compression in VLMs.}
Most VLM efficiency methods shorten the visual sequence produced by the vision encoder.  One line prunes or reorganizes
patch tokens using learned importance signals~\cite{rao2021dynamicvit,liang2022not,alvar2025divprune,shang_llava-prumerge_2024}, and another merges
redundant tokens at inference~\cite{bolya2022token}.  These methods are cheap and effective at moderate compression, but
brittle at very small budgets because a low-scoring token can still contain task-critical evidence; attention-style
importance proxies can also be unstable under aggressive pruning~\cite{jain2019attention,wen2025stop,zhang_beyond_2025-1}.
Learned interfaces instead form compact latent tokens through Perceiver-style resamplers, query modules, or stronger
projectors~\cite{jaegle2021perceiver,alayrac2022flamingo,li2023blip,zhang2025llava,ryoo2021tokenlearner}.
Recent extreme-budget learned-interface methods add query-conditioned aggregation, elastic latent queries, or stronger
projectors~\cite{li2024inference,hu2024matryoshka,li2025tokenpacker}.  They achieve strong accuracy, but add attention-style computation and
alignment complexity to the compression interface.  Braco instead emphasizes a fixed, lightweight visual interface whose
cost can be measured independently of downstream prompting.

\noindent\textbf{Transform-domain compression.}
Transform coding applies structured orthonormal bases so that signal energy concentrates before coefficient selection
~\cite{ahmed1974discrete,wallace1991jpeg,mallat1989multiresolution}.  In VLMs, transform-domain methods such as
Fourier-VLM apply a 2D-DCT, truncate high-frequency coefficients, and reconstruct a coarser spatial grid by inverse DCT before flattening to fewer tokens
~\cite{wang2025fourier,feng2024docpedia}.  Under extreme budgets, however, the retained representation is not only a
signal code but also the visual interface that the LLM must align to.  Braco therefore studies the compressed interface
along two axes: compressibility, determined by basis choice and structured truncation, and learnability, determined by
coordinate organization within the retained subspace.  This view further motivates basis-coordinate embeddings for stable
token identity and a small spatial residual budget for sparse local evidence.

\begin{figure}[htbp]
  \centering
  \includegraphics[width=\linewidth]{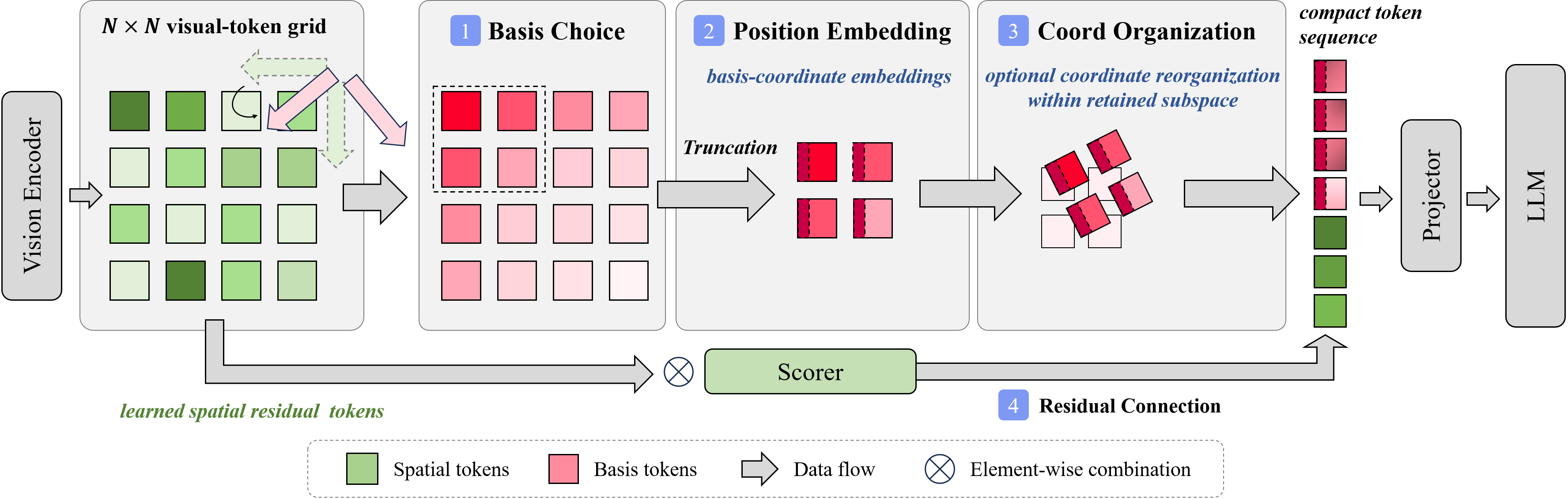}
  \caption{\textbf{Our four-step token coder.}
Starting from an $N\times N$ visual-token grid, Braco transforms tokens into an orthonormal basis, keeps a fixed
$C\times C$ low-frequency block, adds basis-coordinate embeddings, optionally changes coordinates within the same retained
subspace, and appends learned spatial residual tokens pooled in parallel from the original grid.}
  \vspace{-2mm}
  \label{fig:method}
\end{figure}

\section{Methodology}
\label{sec:method}

In this section, we present our token coder and its data path in \cref{fig:method}.  We first introduce our
token parameterization, then formalize a unified view that separates \emph{compressibility} (basis + structured
truncation) from \emph{learnability} (coordinate organization), and specialize these functionals to our design space.
Finally, we instantiate a deployable coder under a strict budget by combining a structured DCT backbone with
basis-coordinate embeddings and a lightweight spatial residual module.

\subsection{Token parameterization}
\label{sec:setup}

Let a vision encoder $f_{\mathrm{vis}}$ produce patch tokens for an image $I$,
\begin{equation}
\mathbf{X}=f_{\mathrm{vis}}(I)\in\mathbb{R}^{L\times D_v}, \qquad L=N^2,
\label{eq:setup_X}
\end{equation}
where $N\times N$ is the patch-grid resolution, $L$ is the number of visual tokens, and $D_v$ is the token dimension.
Let $[L]\triangleq\{1,\dots,L\}$, $\langle\mathbf{A},\mathbf{B}\rangle_F\triangleq\mathrm{tr}(\mathbf{A}^\top\mathbf{B})$,
and let $\mathrm{OffDiag}(\cdot)$ zero diagonal entries.
A token coder outputs $\mathbf{Z}=g(\mathbf{X})\in\mathbb{R}^{K\times D_v}$ before the projector, where $K$ is the token
budget.  Braco parameterizes $g$ by three choices.  First, an orthonormal basis
$\mathbf{B}\in\mathbb{R}^{N\times N}$ with $\mathbf{B}^\top\mathbf{B}=\mathbf{I}_N$ induces
$\mathbf{U}_{\mathbf{B}}=\mathbf{B}\otimes\mathbf{B}\in\mathbb{R}^{L\times L}$ and re-expresses the token lattice as
$\mathbf{Y}=\mathbf{U}_{\mathbf{B}}\mathbf{X}$.  Second, a fixed structured index set
$\mathcal{S}\subseteq[L]$ with $|\mathcal{S}|=K$ keeps $K$ transform coordinates through the row-selection matrix
$\mathbf{P}_{\mathcal{S}}$, giving $\bar{\mathbf{Z}}=\mathbf{P}_{\mathcal{S}}\mathbf{Y}$.  Third, an orthogonal matrix
$\mathbf{A}\in\mathbb{R}^{K\times K}$ may reorganize the retained coordinates:
\begin{equation}
\boxed{
\mathbf{Z}=\mathbf{A}\mathbf{P}_{\mathcal{S}}\mathbf{U}_{\mathbf{B}}\mathbf{X}.
}
\label{eq:unified_general_param}
\end{equation}
Here $(\mathbf{B},\mathcal{S})$ determines \emph{which subspace} is retained, while $\mathbf{A}$ changes only the
\emph{coordinates inside that subspace}.  Varying $(\mathbf{B},\mathcal{S})$ tests information preservation, while varying
$\mathbf{A}$ with $(\mathbf{B},\mathcal{S})$ fixed tests optimization and alignment.  With this parameterization in place,
we next define objectives that assess (1) how much task-relevant information is preserved under a structured retention
rule, and (2) how well the resulting coordinates support downstream optimization.

\subsection{Token compression objectives}
\label{sec:unified_view}

We characterize token compression through two complementary objectives: \textit{compressibility}, the task-relevant
information retained by a small structured set, and \textit{learnability}, the ease of mapping compressed tokens into the
LLM input space.

\paragraph{Compressibility.}
For a dataset $\mathcal{D}$, define the token-axis second moment
$\mathbf{M}=\mathbb{E}_{I_i\sim\mathcal{D}}[\mathbf{X}_i\mathbf{X}_i^\top]$, where
$\mathbf{X}_i=f_{\mathrm{vis}}(I_i)$ is the visual-token grid of sample $I_i$.  The energy retained by a basis--truncation
pair is
\begin{equation}
\mathcal{E}(\mathbf{B};\mathcal{S})
=
\frac{
\mathrm{tr}\!\left(
\mathbf{P}_{\mathcal{S}}\mathbf{U}_{\mathbf{B}}\mathbf{M}\mathbf{U}_{\mathbf{B}}^\top\mathbf{P}_{\mathcal{S}}^\top
\right)}
{\mathrm{tr}(\mathbf{M})}.
\label{eq:unified_energy_retention}
\end{equation}
Here $\mathrm{tr}(\mathbf{M})$ is the total expected token energy; \cref{app:energy_propto} derives this normalized-trace
form.
Because energy may be task-irrelevant and low-energy directions may matter, we also define a readability score
$\mathcal{R}(\mathbf{B};\mathcal{S})\in[0,1]$ as the fraction of a downstream task direction retained by the same subspace.
Let
\begin{equation}
\mathcal{U}(\mathbf{B},\mathcal{S})
\triangleq
\{\mathbf{U}_{\mathbf{B}}^\top\mathbf{P}_{\mathcal{S}}^\top\mathbf{Q}:
\mathbf{Q}\in\mathbb{R}^{K\times D_v}\}
\end{equation}
be the token fields representable from the retained coordinates, and let
$\mathbf{\Pi}_{\mathbf{B},\mathcal{S}}$ be the $\mathbf{M}$-orthogonal projector onto this subspace.  We model the
downstream task locally by a linear target
$f^\star(\mathbf{X})=\langle\mathbf{W}^\star,\mathbf{X}\rangle$, where
$\mathbf{W}^\star\in\mathbb{R}^{L\times D_v}$ is the task direction and $\|\cdot\|_{\mathbf{M}}$ denotes the corresponding
$\mathbf{M}$-weighted norm.  We use
\begin{equation}
\mathcal{R}(\mathbf{B};\mathcal{S})
=
\frac{\mathbb{E}\|\mathbf{\Pi}_{\mathbf{B},\mathcal{S}}\mathbf{W}^\star\|_{\mathbf{M}}^{2}}
{\mathbb{E}\|\mathbf{W}^\star\|_{\mathbf{M}}^{2}},
\end{equation}
with the full construction in \cref{app:readability_term}.  The main compressibility diagnostic is
\begin{equation}
\boxed{
\mathcal{C}(\mathbf{B};\mathcal{S})
=
\lambda\mathcal{E}(\mathbf{B};\mathcal{S})
+
(1-\lambda)\mathcal{R}(\mathbf{B};\mathcal{S}).
}
\label{eq:unified_C}
\end{equation}
where $\lambda\in[0,1]$ trades off energy retention and task-direction retention.
Thus basis choice controls whether a fixed, deployable truncation rule keeps a useful subspace.

\paragraph{Learnability.}
Once $(\mathbf{B},\mathcal{S})$ is fixed, every orthogonal $\mathbf{A}$ in \eqref{eq:unified_general_param} preserves the
same information, but downstream optimization is not invariant to token-axis rotations~\cite{salimans2016weight}.  Let
$\bar{\mathbf{Z}}=\mathbf{P}_{\mathcal{S}}\mathbf{U}_{\mathbf{B}}\mathbf{X}$ and
$\mathbf{G}(\mathbf{A})=\mathbb{E}[(\mathbf{A}\bar{\mathbf{Z}})(\mathbf{A}\bar{\mathbf{Z}})^\top]
=\mathbf{A}\bar{\mathbf{G}}\mathbf{A}^\top$, where
$\bar{\mathbf{G}}=\mathbb{E}[\bar{\mathbf{Z}}\bar{\mathbf{Z}}^\top]\in\mathbb{R}^{K\times K}$ is the Gram matrix before
coordinate organization.  We score $\mathbf{A}$ by a statistical-conditioning penalty
\begin{equation}
\mathcal{L}_{\mathrm{st}}(\mathbf{A})
=
\|\mathrm{OffDiag}(\mathbf{G}(\mathbf{A}))\|_F^2+
\gamma\bigl\|\mathrm{diag}(\mathbf{G}(\mathbf{A}))
-\tfrac{\mathrm{tr}(\bar{\mathbf{G}})}{K}\mathbf{1}\bigr\|_2^2,
\end{equation}
where $\gamma>0$ is a fixed weight, $\mathbf{1}\in\mathbb{R}^{K}$ is the all-ones vector, and
$\mathrm{diag}(\cdot)$ extracts diagonal entries.  \Cref{app:stat_cond_term} gives the expanded Gram-transform view and
term interpretation.
We also use a geometric penalty
\begin{equation}
\mathcal{L}_{\mathrm{geo}}(\mathbf{A})
=
1-\frac{1}{K}\langle\mathbf{A},\mathbf{A}_0\rangle_F
=
\frac{1}{2K}\|\mathbf{A}-\mathbf{A}_0\|_F^2,
\label{eq:unified_geom_obj}
\end{equation}
where $\mathbf{A}_0\in\mathbb{R}^{K\times K}$ is a preferred orthogonal structured organization and
$\langle\mathbf{A},\mathbf{A}_0\rangle_F=\mathrm{tr}(\mathbf{A}^\top\mathbf{A}_0)$.  The combined objective is
\begin{equation}
\boxed{
\mathcal{L}_{\mathrm{learn}}(\mathbf{A};K)
=
\frac{1}{K^2}\mathcal{L}_{\mathrm{st}}(\mathbf{A})
+
\beta\mathcal{L}_{\mathrm{geo}}(\mathbf{A}).
}
\label{eq:unified_learn_obj}
\end{equation}
\Cref{app:final_learn_obj} justifies the budget normalization, and $\beta>0$ controls the trade-off with geometric
compatibility.  Applying this objective to coefficient coordinates and inverse-DCT coarse-grid coordinates gives a
budget-dependent comparison.  Let $\mathbf{I}$ keep coefficient coordinates and let
$\mathbf{U}_C\in\mathbb{R}^{K_b\times K_b}$ be the inverse-DCT coarse-grid organization for the retained $C\times C$
backbone, where $K_b=C^2$.  With
$\Delta_{\mathrm{st}}=\mathcal{L}_{\mathrm{st}}(\mathbf{U}_C)-\mathcal{L}_{\mathrm{st}}(\mathbf{I})$ and
$\rho_C=1-K_b^{-1}\langle\mathbf{I},\mathbf{U}_C\rangle_F$,
\begin{equation}
\mathcal{L}_{\mathrm{learn}}(\mathbf{U}_C;K_b)-\mathcal{L}_{\mathrm{learn}}(\mathbf{I};K_b)
=
\Delta_{\mathrm{st}}/K_b^2-\beta\rho_C .
\end{equation}
Thus coefficient coordinates are favored when statistical conditioning dominates at very small $K_b$, while the coarse
grid becomes preferable once geometric compatibility dominates; \cref{app:coord_candidates} gives the threshold conditions
and random-rotation comparison.

\subsection{Specializing the functionals to our design space}
\label{sec:specialize}

We now specialize the objectives to choose $(\mathbf{B},\mathcal{S},\mathbf{A})$ and motivate the embedding and residual
components.

\paragraph{Basis and structured truncation.}
Step 1 in \cref{fig:method} uses a deployable $C\times C$ low-frequency block $\mathcal{S}_C$ in a separable transform lattice, where $C$ is the
retained frequency cut-off per axis and $K_b=C^2$ is the backbone token count.  This is a standard rule in
transform coding and frequency-domain neural operators~\cite{wallace1991jpeg,dos2020good,qin2021fcanet}.  Among spatial,
DCT, Haar, and random orthonormal bases under this same rule~\cite{ahmed1974discrete,wallace1991jpeg,mallat1989multiresolution,stewart1980efficient},
DCT gives the most favorable empirical combination of energy concentration, task readability, and implementation simplicity in
our diagnostics, so Braco sets $\mathbf{B}=\mathbf{D}_N$, where $\mathbf{D}_N$ is the $N\times N$ orthonormal cosine basis;
\cref{app:choosing_S,app:baselines} give the full comparison.

\paragraph{Position embeddings.}
A structured basis change can improve compressibility but removes explicit spatial identity: after DCT, each retained
token is a global coefficient and token order no longer directly encodes locality.  Braco restores stable indexing cues by
adding the Step 2 input-independent embedding on the retained transform coordinate $(u,v)$; the exact formula is in
\cref{app:basis_coordinate_embedding}.

\paragraph{Coordinate organization.}
For Step 3, with $(\mathbf{B},\mathcal{S})=(\mathbf{D}_N,\mathcal{S}_C)$ fixed, Braco compares three information-equivalent
organizations: coefficient tokens (\texttt{vanilla}), a coarse spatial grid obtained by inverse DCT (\texttt{idct}), and a
random orthogonal rotation (\texttt{randrot}).  The learnability surrogate and experiments agree on a budget-dependent
rule: use coefficient coordinates for very small backbones and switch to the coarse-grid organization once the backbone is
large enough.  In our implementation, $\texttt{vanilla}$ is used when $K_b=C^2<16$, and \texttt{idct} when $K_b\ge16$
(\cref{app:coord_candidates}).

\paragraph{Spatial residual tokens.}
Step 4 in \cref{fig:method} adds a spatial residual branch: the structured backbone carries the compressible global component, but low-pass truncation can still drop
localized evidence.  Braco therefore allocates a residual budget in the spatial domain, where sparse high-frequency details
are naturally concentrated.  The reason is that a spatially sparse residual is diffuse in an incoherent transform basis: for a residual
$\mathbf{r}\in\mathbb{R}^{L}$ with at most $s$ nonzero spatial entries and an orthonormal transform $\mathbf{U}$ with
coherence $\mu=\sqrt{L}\max_{i,j}|U_{ij}|$, the best $m$ transform coefficients satisfy
\begin{equation}
\max_{|\Omega|=m}\|(\mathbf{U}\mathbf{r})_{\Omega}\|_2^2
\le
\frac{m\mu^2s}{L}\|\mathbf{r}\|_2^2 .
\end{equation}
Here $\Omega\subseteq[L]$ indexes the selected transform coordinates.
When $m\mu^2s\ll L$, transform truncation retains little of this localized evidence, motivating Braco's spatial
residual branch; \cref{app:spatial_residual_domain} gives the derivation.

\subsection{The overall framework}
\label{sec:coder}

We summarize our token coder (\cref{fig:method}) and instantiate $g$ under a strict token budget.  It combines (1) a
structured transform-domain backbone, (2) input-independent basis-coordinate embeddings, (3) a budget-dependent orthogonal
coordinate organization within the retained subspace, and (4) a lightweight spatial residual module.  The total budget is
split into a structured backbone and learned residual tokens,
\begin{equation}
K=K_b+K_r,\qquad K_b=C^2,\qquad K_r=S.
\label{eq:coder_budget}
\end{equation}
where $K_b$ is the backbone budget, $K_r$ is the residual-token budget, and $S$ is the number of residual tokens.
Braco applies a 2D DCT to the $N\times N$ visual-token lattice, keeps the $C\times C$ low-frequency block, adds
basis-coordinate embeddings, and applies the budget-dependent coordinate organization above to obtain
$\mathbf{Z}_b\in\mathbb{R}^{C^2\times D_v}$.  In parallel, a lightweight TokenLearner-style scorer produces $S$ sparsemax
weight maps over the original spatial grid~\cite{ryoo2021tokenlearner,martins2016sparsemax}; each residual token is a weighted sum of spatial
tokens,
\begin{equation}
\mathbf{z}^{(s)}_r=(\mathbf{w}^{(s)})^\top\mathbf{X}, \qquad
\mathbf{Z}_r=[\mathbf{z}^{(1)}_r;\dots;\mathbf{z}^{(S)}_r].
\label{eq:coder_pool}
\end{equation}
where $\mathbf{w}^{(s)}$ is the $s$-th sparsemax weight vector over the $L$ spatial locations, $s\in[S]$, and
$\mathbf{z}^{(s)}_r$ is the corresponding residual token.
The final compressed visual interface is $\mathbf{Z}=[\mathbf{Z}_b;\mathbf{Z}_r]$, normalized and projected into the LLM
hidden size.  Residual-pooling and projection details are in \cref{app:coder_details}.  Budget instantiations are in
\cref{app:braco_configs}; stage costs are in \cref{app:module_cost}; training details are in
\cref{app:hardware_protocol}.

\section{Experiments}
We evaluate Braco in two stages.  First, basis and coordinate diagnostics test whether a tiny structured interface retains
readable information and remains learnable, guiding basis selection and coordinate organization.  Second, end-to-end
comparisons under matched token budgets test the resulting empirical accuracy--efficiency frontier, with compressor costs,
larger-input studies, and ablations validating the same design choices.  Full protocols and additional diagnostics are in
\cref{app:extended_experiments}.

\begin{table*}[t]
\caption{\textbf{Main results under matched visual-token budgets.}
We report eight benchmark scores, Acc. ($\uparrow$; the mean score normalized by the 576-token Vanilla model), and
single-image full-pipeline prefill FLOPs/latency ($\downarrow$).
QueCC is omitted at 25 tokens because its native grid does not support a $5\times5$ output on the fixed $24\times24$
visual-token lattice.
Braco is on the leading empirical Acc.--cost frontier at 25/16/9 tokens and remains within 0.2 Acc. of QueCC at 4 tokens with lower cost.}
\centering
\small
\renewcommand{\twoline}[2]{\fontsize{8.8}{9.0}\selectfont #1}
\setlength{\tabcolsep}{2.3pt}
\setlength{\extrarowheight}{0.4pt}
\renewcommand{\arraystretch}{1.0}
\resizebox{\textwidth}{!}{%
\newcommand{\vmidrule}{}
\begin{tabular}{@{}l|*{8}{c}|c|c|c@{}}
\toprule
\textbf{Method} &
\textbf{GQA} & \textbf{MMB$^{\text{EN}}$} & \textbf{MMB$^{\text{CN}}$} & \textbf{MME$^{\text{All}}$} & \textbf{POPE$^{\text{F1}}$} & \textbf{SQA} & \textbf{VQA-T} & \textbf{MMVet} &
\makecell[c]{\textbf{Acc.}\\[-1pt]\textbf{(\%)}} & \makecell[c]{\textbf{FLOPs}\\[-1pt]\textbf{(T)}} & \makecell[c]{\textbf{Lat.}\\[-1pt]\textbf{(ms)}} \\
\specialrule{\lightrulewidth}{0pt}{0pt}

\rowcolor{gray!15}
\multicolumn{12}{c}{\textit{Upper Bound, 576 Tokens} (\textbf{1$\times$})} \\
Vanilla &
\twoline{62.9}{100.0\%} & \twoline{65.5}{100.0\%} & \twoline{60.7}{100.0\%} & \twoline{1785}{100.0\%} &
\twoline{85.7}{100.0\%} & \twoline{69.7}{100.0\%} & \twoline{58.0}{100.0\%} & \twoline{32.8}{100.0\%} &
100.0 & 8.67 & 67.25 \\
\specialrule{\lightrulewidth}{0pt}{0pt}

\rowcolor{gray!15}
\multicolumn{12}{c}{\textit{25 Retained Tokens} (\textbf{23$\times$})} \\
PruMerge {\scriptsize (ICCV25)} &
\twoline{49.8}{79.2\%} & \twoline{55.8}{85.2\%} & \twoline{47.0}{77.4\%} & \twoline{1515}{84.9\%} &
\twoline{58.5}{68.3\%} & \twoline{68.7}{98.6\%} & \twoline{50.4}{86.9\%} & \twoline{20.2}{61.6\%} &
80.2 & 1.37 & 44.10 \\
\vmidrule
DivPrune {\scriptsize (CVPR25)} &
\twoline{54.0}{85.9\%} & \twoline{59.8}{91.3\%} & \twoline{50.8}{83.7\%} & \twoline{1510}{84.6\%} &
\twoline{75.2}{87.7\%} & \twoline{68.5}{98.3\%} & \twoline{49.5}{85.3\%} & \twoline{26.0}{79.3\%} &
87.0 & 1.40 & 41.00 \\
\vmidrule
MQT-LLaVA {\scriptsize (NIPS24)} &
\twoline{57.1}{90.8\%} & \twoline{61.4}{93.7\%} & \twoline{53.1}{87.5\%} & \twoline{1689}{94.6\%} &
\twoline{79.9}{93.2\%} & \twoline{69.4}{99.6\%} & \twoline{50.2}{86.6\%} & \twoline{27.7}{84.5\%} &
91.3 & 1.37 & 44.20 \\
\vmidrule
TokenPacker {\scriptsize (IJCV25)} &
\twoline{57.4}{91.3\%} & \twoline{64.1}{97.9\%} & \twoline{55.8}{91.9\%} & \twoline{1688}{94.6\%} &
\twoline{\textbf{83.0}}{96.8\%} & \twoline{69.4}{99.6\%} & \twoline{\textbf{53.3}}{91.9\%} & \twoline{29.3}{89.3\%} &
94.2 & 1.38 & 38.40 \\
\vmidrule
Fourier-VLM &
\twoline{\textbf{58.4}}{92.8\%} & \twoline{63.3}{96.6\%} & \twoline{54.6}{90.0\%} & \twoline{\textbf{1730}}{96.9\%} &
\twoline{\textbf{83.0}}{96.8\%} & \twoline{69.7}{100.0\%} & \twoline{50.2}{86.6\%} & \twoline{27.5}{83.8\%} &
92.9 & 1.37 & 39.83 \\
\vmidrule
\textbf{Braco (ours)} &
\twoline{58.1}{92.4\%} & \twoline{\textbf{65.4}}{99.8\%} & \twoline{\textbf{57.2}}{94.2\%} & \twoline{1705}{95.5\%} &
\twoline{82.6}{96.4\%} & \twoline{\textbf{70.0}}{100.4\%} & \twoline{52.3}{90.2\%} & \twoline{\textbf{30.4}}{92.7\%} &
\textbf{95.2} & 1.37 & 41.03 \\
\specialrule{\lightrulewidth}{0pt}{0pt}

\rowcolor{gray!15}
\multicolumn{12}{c}{\textit{16 Retained Tokens} (\textbf{36$\times$})} \\
PruMerge {\scriptsize (ICCV25)} &
\twoline{46.9}{74.6\%} & \twoline{53.1}{81.1\%} & \twoline{44.2}{72.8\%} & \twoline{1446}{81.0\%} &
\twoline{51.6}{60.2\%} & \twoline{68.2}{97.8\%} & \twoline{49.3}{85.0\%} & \twoline{18.8}{57.3\%} &
76.2 & 1.25 & 43.37 \\
\vmidrule
DivPrune {\scriptsize (CVPR25)} &
\twoline{52.6}{83.6\%} & \twoline{56.7}{86.6\%} & \twoline{47.4}{78.1\%} & \twoline{1411}{79.0\%} &
\twoline{71.3}{83.2\%} & \twoline{67.3}{96.6\%} & \twoline{48.3}{83.3\%} & \twoline{24.8}{75.6\%} &
83.2 & 1.28 & 40.26 \\
\vmidrule
MQT-LLaVA {\scriptsize (NIPS24)} &
\twoline{55.3}{87.9\%} & \twoline{62.8}{95.9\%} & \twoline{53.9}{88.8\%} & \twoline{1630}{91.3\%} &
\twoline{78.2}{91.2\%} & \twoline{68.9}{98.9\%} & \twoline{48.2}{83.1\%} & \twoline{27.6}{84.1\%} &
90.2 & 1.25 & 44.27 \\
\vmidrule
QueCC {\scriptsize (ICLR25)} &
\twoline{\textbf{59.0}}{93.8\%} & \twoline{63.1}{96.3\%} & \twoline{54.6}{90.0\%} & \twoline{1668}{93.4\%} &
\twoline{\textbf{83.5}}{97.4\%} & \twoline{\textbf{70.6}}{101.3\%} & \twoline{52.8}{91.0\%} & \twoline{28.8}{87.8\%} &
93.9 & 1.36 & 63.22 \\
\vmidrule
TokenPacker {\scriptsize (IJCV25)} &
\twoline{57.0}{90.6\%} & \twoline{\textbf{63.7}}{97.3\%} & \twoline{55.2}{90.9\%} & \twoline{1681}{94.2\%} &
\twoline{83.0}{96.8\%} & \twoline{69.3}{99.4\%} & \twoline{\textbf{53.4}}{92.1\%} & \twoline{29.0}{88.4\%} &
93.7 & 1.26 & 37.87 \\
\vmidrule
Fourier-VLM &
\twoline{56.7}{90.1\%} & \twoline{61.4}{93.7\%} & \twoline{51.0}{84.0\%} & \twoline{1644}{92.1\%} &
\twoline{82.5}{96.3\%} & \twoline{69.6}{99.9\%} & \twoline{48.5}{83.6\%} & \twoline{27.2}{82.9\%} &
90.3 & 1.25 & 40.00 \\
\vmidrule
\textbf{Braco (ours)} &
\twoline{57.5}{91.4\%} & \twoline{63.3}{96.6\%} & \twoline{\textbf{55.8}}{91.9\%} & \twoline{\textbf{1704}}{95.5\%} &
\twoline{82.4}{96.1\%} & \twoline{69.5}{99.7\%} & \twoline{51.9}{89.5\%} & \twoline{\textbf{29.8}}{90.9\%} &
\textbf{94.0} & 1.25 & 40.59 \\
\specialrule{\lightrulewidth}{0pt}{0pt}

\rowcolor{gray!15}
\multicolumn{12}{c}{\textit{9 Retained Tokens} (\textbf{64$\times$})} \\
PruMerge {\scriptsize (ICCV25)} &
\twoline{44.5}{70.7\%} & \twoline{48.8}{74.5\%} & \twoline{40.6}{66.9\%} & \twoline{1352}{75.7\%} &
\twoline{47.0}{54.8\%} & \twoline{66.5}{95.4\%} & \twoline{46.1}{79.5\%} & \twoline{15.8}{48.2\%} &
70.7 & 1.15 & 42.90 \\
\vmidrule
DivPrune {\scriptsize (CVPR25)} &
\twoline{50.1}{79.7\%} & \twoline{53.5}{81.7\%} & \twoline{44.2}{72.8\%} & \twoline{1320}{73.9\%} &
\twoline{67.8}{79.1\%} & \twoline{65.8}{94.4\%} & \twoline{46.2}{79.7\%} & \twoline{22.0}{67.1\%} &
78.5 & 1.17 & 39.91 \\
\vmidrule
MQT-LLaVA {\scriptsize (NIPS24)} &
\twoline{53.9}{85.7\%} & \twoline{62.4}{95.3\%} & \twoline{52.9}{87.1\%} & \twoline{1565}{87.7\%} &
\twoline{78.8}{91.9\%} & \twoline{69.5}{99.7\%} & \twoline{47.1}{81.2\%} & \twoline{27.1}{82.6\%} &
88.9 & 1.15 & 43.02 \\
\vmidrule
QueCC {\scriptsize (ICLR25)} &
\twoline{\textbf{58.3}}{92.7\%} & \twoline{62.9}{96\%} & \twoline{\textbf{55.6}}{91.6\%} & \twoline{1707}{95.6\%} &
\twoline{\textbf{83.3}}{97.2\%} & \twoline{69.0}{99.0\%} & \twoline{\textbf{51.4}}{88.6\%} & \twoline{27.6}{84.1\%} &
93.1 & 1.26 & 62.79 \\
\vmidrule
TokenPacker {\scriptsize (IJCV25)} &
\twoline{55.9}{88.9\%} & \twoline{62.5}{95.4\%} & \twoline{53.5}{88.1\%} & \twoline{1645}{92.2\%} &
\twoline{81.8}{95.4\%} & \twoline{68.7}{98.6\%} & \twoline{51.2}{88.3\%} & \twoline{27.7}{84.5\%} &
91.4 & 1.16 & 37.40 \\
\vmidrule
Fourier-VLM &
\twoline{56.4}{89.7\%} & \twoline{60.5}{92.4\%} & \twoline{50.4}{83.0\%} & \twoline{1683}{94.3\%} &
\twoline{81.4}{95.0\%} & \twoline{67.4}{96.7\%} & \twoline{47.0}{81.0\%} & \twoline{24.9}{75.9\%} &
88.5 & 1.15 & 39.88 \\
\vmidrule
\textbf{Braco (ours)} &
\twoline{57.0}{90.6\%} & \twoline{\textbf{63.4}}{96.8\%} & \twoline{54.2}{89.3\%} & \twoline{\textbf{1773}}{99.3\%} &
\twoline{81.7}{95.3\%} & \twoline{\textbf{69.7}}{100.0\%} & \twoline{51.1}{88.1\%} & \twoline{\textbf{28.2}}{86.0\%} &
\textbf{93.2} & 1.15 & 40.22 \\
\specialrule{\lightrulewidth}{0pt}{0pt}

\rowcolor{gray!15}
\multicolumn{12}{c}{\textit{4 Retained Tokens} (\textbf{144$\times$})} \\
PruMerge {\scriptsize (ICCV25)} &
\twoline{39.0}{62.0\%} & \twoline{43.1}{65.8\%} & \twoline{35.4}{58.3\%} & \twoline{1210}{67.8\%} &
\twoline{38.5}{44.9\%} & \twoline{62.9}{90.2\%} & \twoline{40.2}{69.3\%} & \twoline{12.3}{37.5\%} &
62.0 & 1.09 & 42.60 \\
\vmidrule
DivPrune {\scriptsize (CVPR25)} &
\twoline{44.2}{70.3\%} & \twoline{47.9}{73.1\%} & \twoline{39.6}{65.2\%} & \twoline{1199}{67.2\%} &
\twoline{60.0}{70.0\%} & \twoline{62.6}{89.8\%} & \twoline{40.8}{70.3\%} & \twoline{18.2}{55.5\%} &
70.2 & 1.11 & 39.50 \\
\vmidrule
MQT-LLaVA {\scriptsize (NIPS24)} &
\twoline{51.5}{81.9\%} & \twoline{\textbf{62.9}}{96.0\%} & \twoline{53.4}{88.0\%} & \twoline{1449}{81.2\%} &
\twoline{78.2}{91.2\%} & \twoline{69.8}{100.1\%} & \twoline{46.2}{79.7\%} & \twoline{25.8}{78.7\%} &
87.1 & 1.09 & 44.82 \\
\vmidrule
QueCC {\scriptsize (ICLR25)} &
\twoline{\textbf{56.5}}{89.8\%} & \twoline{61.0}{93.1\%} & \twoline{53.5}{88.1\%} & \twoline{1641}{91.9\%} &
\twoline{\textbf{81.7}}{95.3\%} & \twoline{68.2}{97.8\%} & \twoline{\textbf{50.4}}{86.9\%} & \twoline{\textbf{28.8}}{87.8\%} &
\textbf{91.4} & 1.20 & 64.17 \\
\vmidrule
TokenPacker {\scriptsize (IJCV25)} &
\twoline{53.0}{84.3\%} & \twoline{60.2}{91.9\%} & \twoline{51.0}{84.0\%} & \twoline{1550}{86.8\%} &
\twoline{79.0}{92.2\%} & \twoline{66.8}{95.8\%} & \twoline{47.0}{81.0\%} & \twoline{24.0}{73.2\%} &
86.2 & 1.10 & 37.10 \\
\vmidrule
Fourier-VLM &
\twoline{54.0}{85.9\%} & \twoline{56.1}{85.6\%} & \twoline{45.4}{74.8\%} & \twoline{1609}{90.1\%} &
\twoline{81.0}{94.5\%} & \twoline{64.8}{93\%} & \twoline{45.7}{78.8\%} & \twoline{21.5}{65.5\%} &
83.5 & 1.09 & 40.30 \\
\vmidrule
\textbf{Braco (ours)} &
\twoline{54.4}{86.5\%} & \twoline{62.2}{95\%} & \twoline{\textbf{53.8}}{88.6\%} & \twoline{\textbf{1658}}{92.9\%} &
\twoline{80.7}{94.2\%} & \twoline{\textbf{69.9}}{100.3\%} & \twoline{50.2}{86.6\%} & \twoline{28.0}{85.4\%} &
91.2 & 1.09 & 40.97 \\
\bottomrule
\end{tabular}%
}
\vspace{-3mm}
\label{tab:token-retain}
\end{table*}

\begin{figure}[b]
  \centering
  \vspace{-5mm}
  \newlength{\figthreeheight}
  \setlength{\figthreeheight}{2.75cm}
  \begin{subfigure}[t]{0.315\textwidth}
    \centering
    \includegraphics[height=\figthreeheight]{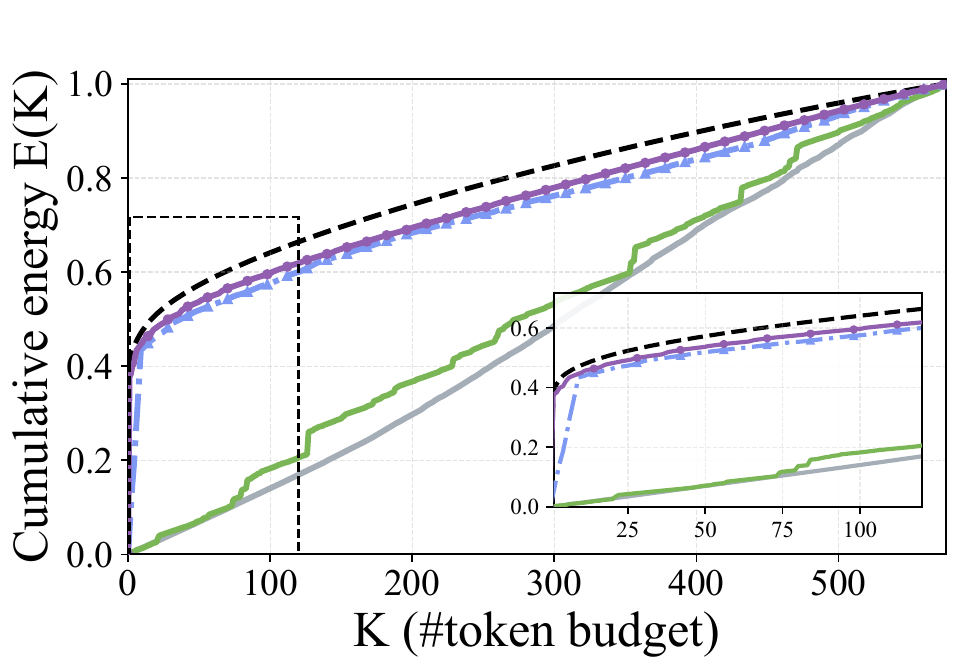}
    \caption{Structured truncation}
    \label{fig:energy_curves}
  \end{subfigure}\hfill
  \begin{subfigure}[t]{0.315\textwidth}
    \centering
    \includegraphics[height=\figthreeheight]{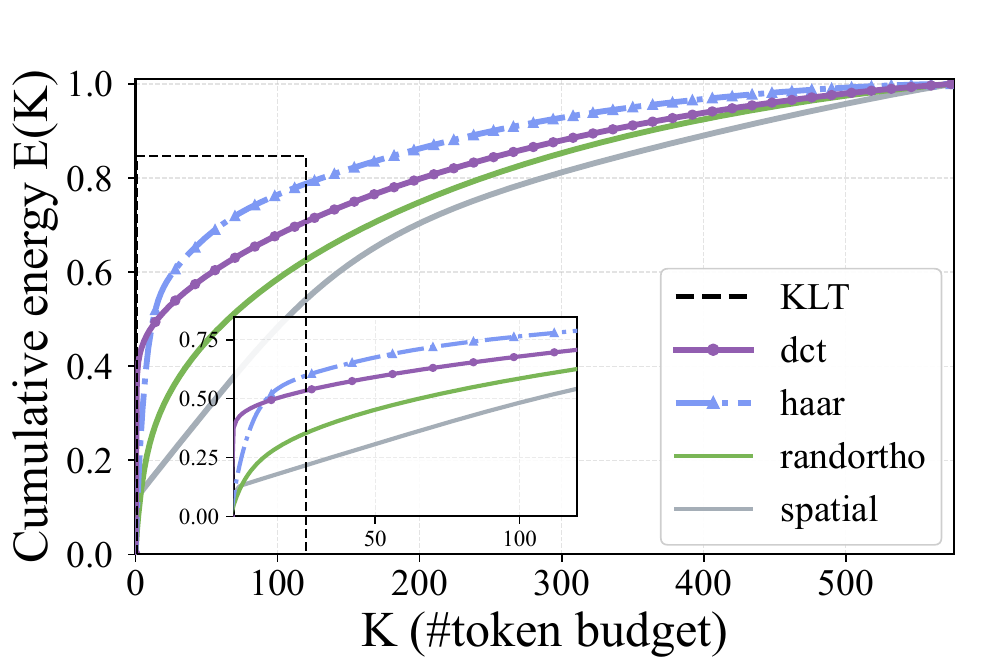}
    \caption{Magnitude truncation}
    \label{fig:energy_curves_magnitude}
  \end{subfigure}\hfill
  \begin{subfigure}[t]{0.34\textwidth}
    \centering
    \includegraphics[height=\figthreeheight]{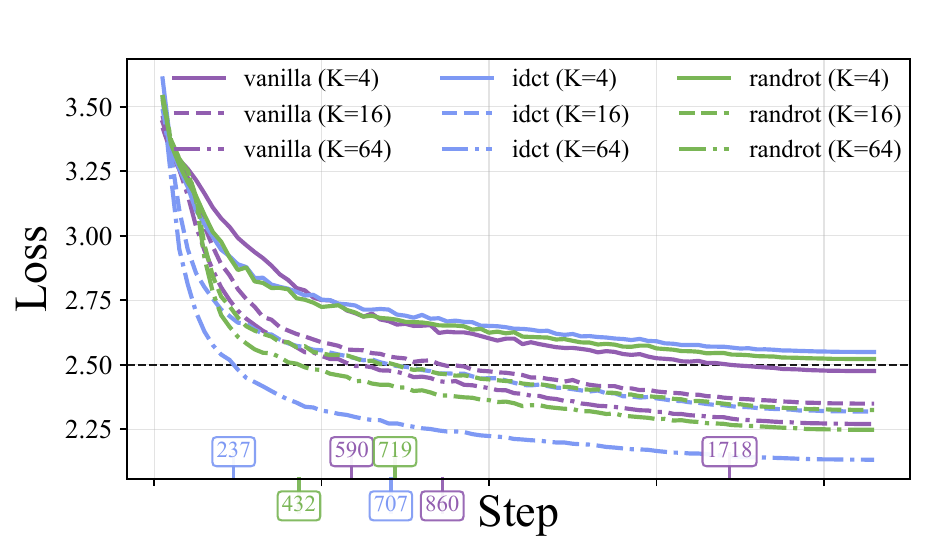}
    \caption{Development loss}
    \label{fig:learnability_loss_curves_dev}
  \end{subfigure}
  \vspace{-1mm}
  \caption{\textbf{Compression vs.\ optimization probes.}
  (a--b) Orthonormal basis choice controls energy retention under deployable structured truncation and oracle magnitude truncation.
  KLT is an oracle upper bound fitted to the diagnostic token second moment.
  (c) With an identical retained subspace, coordinate organization can still change optimization behavior.}
  \label{fig:exp_compact_vs_learn}
\end{figure}

% ============================================================
\subsection{Compressibility under basis choice and structured truncation}
\label{sec:exp_compressibility}

\begin{figure}[t]
  \centering
  \captionsetup{skip=2pt}
  \captionsetup[subfigure]{skip=1pt}
  \newlength{\figfourheight}
  \newlength{\energymapwidth}
  \setlength{\figfourheight}{2.52cm}
  \begin{subfigure}[c]{0.39\textwidth}
    \centering
    \begin{minipage}[c][\figfourheight][c]{\linewidth}
      \centering
      \setlength{\energymapwidth}{0.242\linewidth}%
      \includegraphics[width=\energymapwidth]{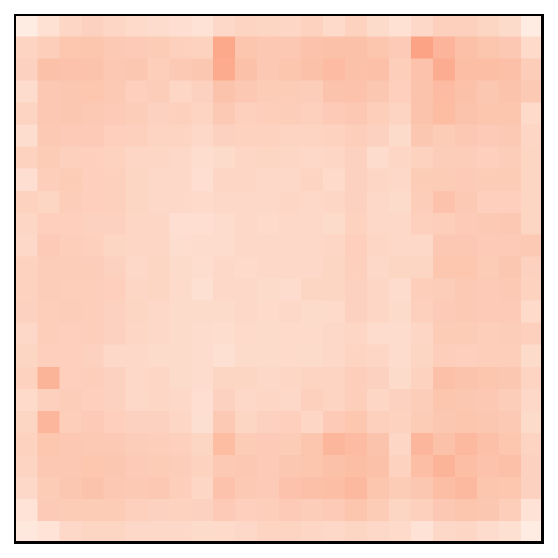}\hspace{0.006\linewidth}%
      \includegraphics[width=\energymapwidth]{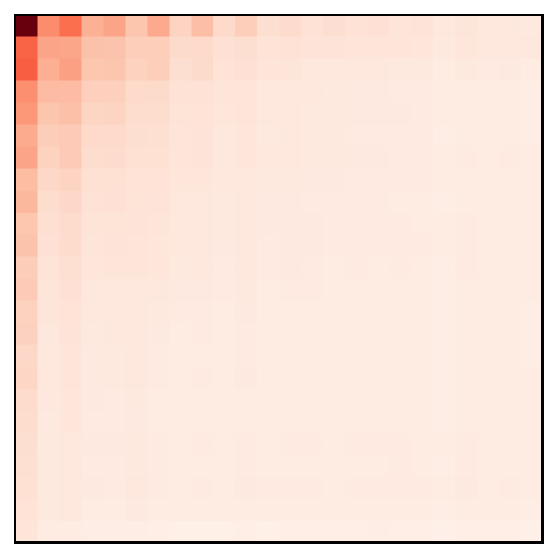}\hspace{0.006\linewidth}%
      \includegraphics[width=\energymapwidth]{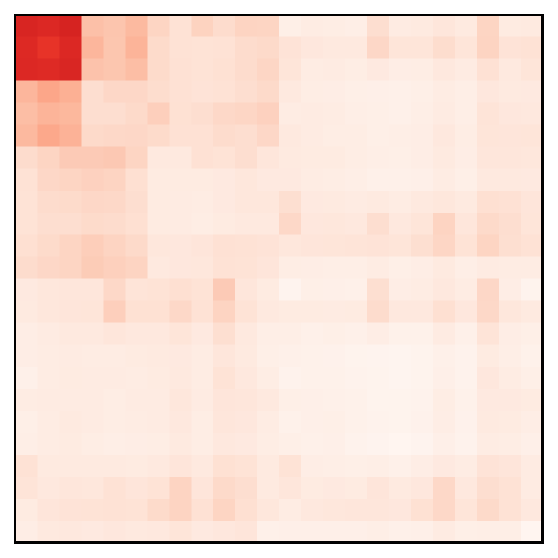}\hspace{0.006\linewidth}%
      \includegraphics[width=\energymapwidth]{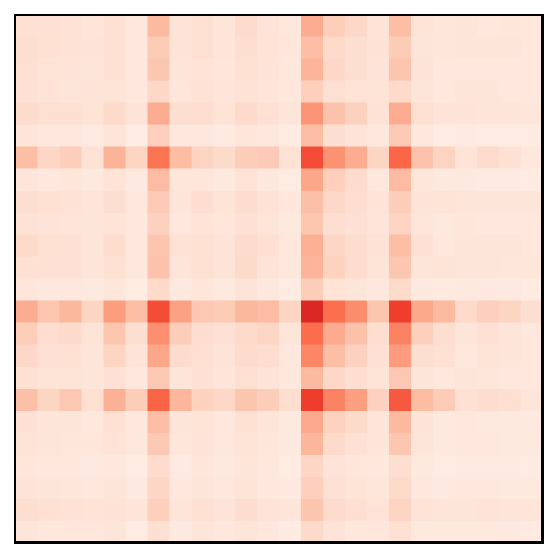}\\[1pt]
      \makebox[\energymapwidth][c]{\scriptsize\bfseries Spatial}\hspace{0.006\linewidth}%
      \makebox[\energymapwidth][c]{\scriptsize\bfseries DCT}\hspace{0.006\linewidth}%
      \makebox[\energymapwidth][c]{\scriptsize\bfseries Haar}\hspace{0.006\linewidth}%
      \makebox[\energymapwidth][c]{\scriptsize\bfseries RandOrtho}
    \end{minipage}
    \caption{Energy}
    \label{fig:exp_energy_maps}
  \end{subfigure}\hfill
  \begin{subfigure}[c]{0.285\textwidth}
    \centering
    \includegraphics[height=\figfourheight,width=\linewidth]{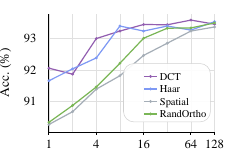}
    \caption{Validation split}
    \label{fig:exp_linear_probe_val}
  \end{subfigure}\hfill
  \begin{subfigure}[c]{0.285\textwidth}
    \centering
    \includegraphics[height=\figfourheight,width=\linewidth]{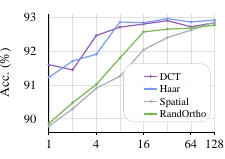}
    \caption{Test split}
    \label{fig:exp_linear_probe_test}
  \end{subfigure}
  \vspace{1mm}
  \caption{\textbf{Compact compressibility diagnostics.}
  (a) DCT concentrates energy into a compact low-frequency region.
  (b--c) CelebA linear probes test task-readability under the same structured truncation rule.}
  \vspace{-4mm}
  \label{fig:exp_compressibility_diag}
\end{figure}

Under an input-independent, deployable truncation rule, we test whether basis choice determines how much task evidence
survives at tiny token budgets.  Before multimodal training, we measure two properties of the retained coordinates:
generic token-field energy and linearly readable semantic directions.  We
report energy retention $E(K)$, the empirical counterpart of
$\mathcal{E}(\mathbf{B};\mathcal{S})$ in \cref{eq:unified_energy_retention} at budget $K$, comparing spatial, DCT, Haar,
random orthonormal, and KLT-oracle bases under structured truncation and magnitude truncation; KLT serves as a fitted
oracle reference for this diagnostic.
DCT/Haar retain far more energy than spatial or random bases (\cref{fig:energy_curves}), e.g.,
$\approx0.51$ vs.\ $\approx0.04$ at $K{=}32$ and $\approx0.57$ vs.\ $\approx0.08$ at $K{=}64$; the gap largely disappears
under magnitude truncation (\cref{fig:energy_curves_magnitude}), showing that the fixed deployable ordering matters.  Separately, the CelebA~\cite{liu2015faceattributes} probes in
\cref{fig:exp_linear_probe_val,fig:exp_linear_probe_test} measure task-readability under structured truncation: using
frozen patch tokens and the same linear-probe setup, DCT reaches
$91.6\%$ vs.\ $89.8\%$ for spatial at $K{=}1$ and $92.5\%$ vs.\ $90.9\%$ at $K{=}4$.  Together with the maps, protocol
details, and full probe curves in \cref{fig:exp_energy_maps,app:compression_diagnostics}, these diagnostics
identify basis choice as an important lever for making a fixed tiny interface informative.

% ============================================================
\captionsetup[table]{skip=5pt}
\captionsetup[subtable]{skip=4pt}

\subsection{Learnability under subspace-preserving coordinate organizations}
\label{sec:exp_learnability}

The learnability diagnostic isolates a different question from compressibility: even with an identical retained subspace,
coordinate organization can change optimization.  We fix the same low-frequency DCT backbone subspace and vary only its coordinates: coefficient
tokens (\texttt{vanilla}), a coarse spatial grid (\texttt{idct}), or a random orthogonal rotation.  Since the variants span
the same subspace, we run only the pretraining objective, split the stream 0.99/0.01 into train/dev, and use
time-to-threshold on held-out dev cross-entropy ($<2.50$) to isolate optimization effects.  The pattern is budget-dependent
(\cref{fig:learnability_loss_curves_dev}): at $K_b{=}4$, only \texttt{vanilla} reaches the target; at $K_b{=}16$ and
$K_b{=}64$, \texttt{idct} reaches it 18\% and 60\% faster.  This motivates coordinate
organization as a separate design axis in \cref{sec:unified_view}; the objective-level derivation and candidates are given
in \cref{app:final_learn_obj,app:coord_candidates}.

\subsection{End-to-end comparison}
\label{sec:exp_main}

\begin{table}[!b]
  \centering
  \vspace{-4mm}
  \caption{\textbf{Compressor cost at 16 tokens.}
  Boundary denotes pre- or post-projector measurement.}
  \label{tab:main_compressor_cost}
  \small
  \setlength{\tabcolsep}{2.3pt}
  \renewcommand{\arraystretch}{1.04}
  \begin{tabular}{@{}llcccc@{}}
    \toprule
    \textbf{Method} & \textbf{Boundary} & \textbf{Acc. (\%)} & \textbf{Lat. (ms)} & \textbf{FLOPs (G)} & \textbf{Mem. (MB)} \\
    \midrule
    QueCC {\scriptsize (ICLR25)} & pre-proj. & 93.9 & 17.864 & 109.504 & 14.875 \\
    MQT-LLaVA {\scriptsize (NIPS24)} & pre-proj. & 90.2 & 5.154 & 2.674 & 11.336 \\
    PruMerge {\scriptsize (ICCV25)} & pre-proj. & 76.2 & 3.300 & 0.724 & 13.749 \\
    \textbf{Braco (ours)} & pre-proj. & 94.0 & 1.073 & 1.389 & 8.731 \\
    \midrule
    TokenPacker {\scriptsize (IJCV25)} & post-proj. & 93.7 & 1.004 & 15.271 & 15.125 \\
    DivPrune {\scriptsize (CVPR25)} & post-proj. & 83.2 & 1.061 & 29.603 & 23.008 \\
    \textbf{Braco (ours)} & post-proj. & 94.0 & 1.270 & 2.060 & 8.731 \\
    \bottomrule
  \end{tabular}
\end{table}

We now evaluate Braco end-to-end under \emph{hard, deployment-friendly} visual-token budgets and compare it to prior token
compression paradigms under matched budgets.

\paragraph{Benchmarks and metrics.}
We evaluate on GQA~\cite{hudson2019gqa}, MMBench (EN/CN)~\cite{liu2024mmbench},
MME (All)~\cite{fu2023mme}, POPE (F1)~\cite{li2023pope}, ScienceQA~\cite{lu2022scienceqa},
VQA-Text (TextVQA)~\cite{singh2019textvqa}, and MMVet~\cite{yu2024mmvet}; \cref{tab:token-retain}
reports per-benchmark scores, Vanilla-normalized Acc., and single-image full-pipeline prefill FLOPs/latency from the vision
encoder through LLM prefill.
All baselines share the retraining/evaluation harness and matched budgets.  \Cref{app:hardware_protocol,tab:app_training_seeds}
give the Acc. definition, training/measurement protocols, and three-seed results.

\paragraph{Baselines and implementation.}
Baselines cover pruning/merging (PruMerge~\cite{shang_llava-prumerge_2024}, DivPrune~\cite{alvar2025divprune}),
learned interfaces (MQT-LLaVA~\cite{hu2024matryoshka}, QueCC~\cite{li2024inference}, TokenPacker~\cite{li2025tokenpacker}),
and transform coding (Fourier-VLM~\cite{wang2025fourier}), all evaluated under matched retained-token budgets.  We implement
\textbf{Braco} on \textbf{LLaVA-1.5-7B}~\cite{liu2024improved} by replacing the dense visual stream with a structured
backbone plus spatial residual interface.  For each budget, Braco splits tokens between a low-frequency backbone and
spatial residuals: $c1s3/c2s5/c3s7/c4s9$ for $K=4/9/16/25$.  Exact coordinate choices are in
\cref{app:braco_configs}.

\paragraph{Main results.}
\Cref{tab:token-retain} shows that Braco provides a favorable accuracy--efficiency trade-off under matched extreme token budgets.
Relative to the 576-token Vanilla model, Braco reduces full-pipeline prefill FLOPs from 8.67T to 1.09--1.37T while retaining
91.2--95.2 Vanilla-normalized Acc. across 4--25 tokens.  Compared with prior compression methods, Braco is consistently
competitive in aggregate accuracy at similar or lower compute: it attains the highest Acc. among the evaluated methods at
25, 16, and 9 tokens, and remains within 0.2 Acc. of QueCC at 4 tokens.  The comparison with QueCC highlights the main
efficiency difference: at 16 and 9 tokens, Braco matches QueCC within 0.1 Acc. while reducing latency by about 36\%; at
4 tokens, it keeps a similar aggregate score with lower FLOPs and substantially lower latency.  Together, these results
show that Braco preserves much of Vanilla's aggregate performance at substantially lower prefill compute and latency.

\paragraph{Compressor cost.}
We further isolate the 16-token compressor to measure the cost of the compression module itself.
Braco matches QueCC's accuracy with 16.6$\times$ lower module latency and
78.8$\times$ fewer compressor FLOPs (\cref{tab:main_compressor_cost}). 
The pre-/post-projector split rules out a boundary artifact: Braco keeps QueCC-level accuracy pre-projector and far lower post-projector FLOPs than TokenPacker/DivPrune.
Stage and multi-budget costs are in \cref{app:module_cost}.

\paragraph{Generalization.}
Beyond the 576-token Vicuna-7B~\cite{vicuna2023} setting, \cref{tab:main_larger_visual_tokens} shows that Braco keeps 98.1 Acc. at 2880 input tokens with Vicuna-7B.
The 576-token sweep primarily stresses whether accuracy survives extreme compression, since full-pipeline FLOPs and latency in this regime also include token-insensitive vision and prompt-processing cost.
As the visual interface grows, however, keeping the retained-token budget small turns extreme compression into a larger end-to-end saving: Braco removes a much larger uncompressed prefill burden, reducing full-pipeline FLOPs from 40.57T to 3.45T and latency from 261.08ms to 44.36ms at 2880 input tokens.
With Qwen2.5-3B~\cite{yang2024qwen25technicalreport}, Braco retains 93.1/92.6 Acc. at 729/1024 input tokens and
90.8 Acc. at 3645 input tokens ($182\times$ compression).  QueCC is close in the 576-token Vicuna setting but degrades on
Qwen2.5-3B, where its query/downsampling overhead is less well amortized.
This may also reflect the sensitivity of query-dependent compression to weaker prompt understanding in smaller LLMs.
Braco remains prompt-independent and becomes more valuable as uncompressed prefill cost grows.  Full inference and training-time breakdowns are in
\cref{app:larger_token_generalization}.

\subsection{Ablations}
\label{sec:exp_ablation}

We ablate Braco's \emph{structured backbone + spatial residual} interface using the same Vanilla-normalized Acc. as in \cref{tab:token-retain}.

\begin{figure}[t]
  \centering
  \begin{subfigure}[t]{0.32\textwidth}
    \centering
    \includegraphics[width=\linewidth]{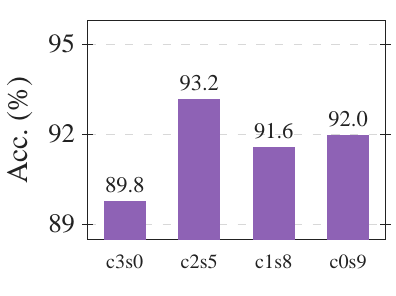}
    \caption{Allocation ($K=9$)}
    \label{fig:exp_ablation_cnsn}
  \end{subfigure}\hfill
  \begin{subfigure}[t]{0.32\textwidth}
    \centering
    \includegraphics[width=\linewidth]{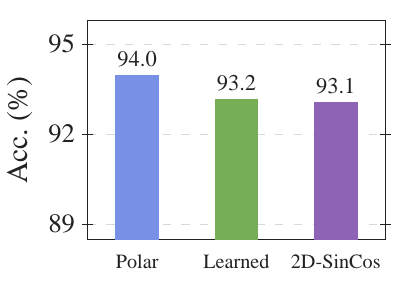}
    \caption{PE variants ($K=16$)}
    \label{fig:exp_ablation_pe_alt}
  \end{subfigure}\hfill
  \begin{subfigure}[t]{0.345\textwidth}
    \centering
    \includegraphics[width=\linewidth]{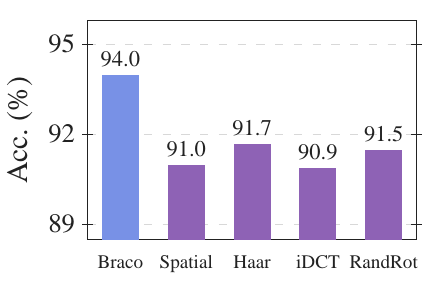}
    \caption{Substitutions ($K=16$)}
    \label{fig:exp_ablation_step}
  \end{subfigure}
  \caption{\textbf{Ablations.}
  (a) Backbone--residual allocation under the 9-token budget.
  (b) Basis-coordinate embedding variants under the 16-token budget.
  (c) Basis and coordinate substitutions under the 16-token budget.}
  \vspace{-4mm}
  \label{fig:exp_ablation}
\end{figure}

At 9 tokens, hybrid \texttt{c2s5} reaches 93.2 Acc. (\cref{fig:exp_ablation_cnsn}), above pure-backbone \texttt{c3s0} (89.8) and residual-only
\texttt{c0s9} (92.0) at comparable compressor cost, indicating that hybrid allocation uses the small token budget more
effectively than either branch alone.
At the 16-token \texttt{c3s7} setting, replacing the DCT backbone with spatial/Haar tokens costs 2.3--3.0 Acc. points,
replacing the selected coordinate organization with \texttt{idct}/random rotation costs 2.5--3.1 points
(\cref{fig:exp_ablation_step}), and replacing Polar Fourier embeddings with learned/2D sine--cosine variants costs 0.8--0.9
points (\cref{fig:exp_ablation_pe_alt}).
Additional costs and design-space definitions are in
\cref{app:module_cost,app:basis_coordinate_embedding,app:coord_candidates,app:spatial_residual_domain}.

\section{Conclusion}
We propose a deployable token coder for extreme visual-token compression in vision-encoder$\rightarrow$LLM pipelines.
The design disentangles \emph{compressibility} (basis transform + structured truncation) from \emph{learnability}
(coordinate organization within the retained subspace), enabling principled comparisons under a fixed interface.
Instantiated with a structured backbone, basis-coordinate embeddings, budget-dependent coordinate organization, and a
lightweight sparse-pooled spatial residual, Braco delivers a strong accuracy--efficiency trade-off.

\clearpage
\begin{ack}
This work received no external funding. The authors declare no competing interests.
\end{ack}

{\small
\bibliographystyle{plainnat}
\bibliography{neurips_2026_refs}

@article{ahmed1974discrete,
  title={Discrete Cosine Transform},
  author={Ahmed, Nasir and Natarajan, T. and Rao, Kamisetty R.},
  journal={IEEE Transactions on Computers},
  volume={C-23},
  number={1},
  pages={90--93},
  year={1974},
  doi={10.1109/T-C.1974.223784}
}

@article{alayrac2022flamingo,
  title={Flamingo: A Visual Language Model for Few-Shot Learning},
  author={Alayrac, Jean-Baptiste and Donahue, Jeff and Luc, Pauline and Miech, Antoine and Barr, Iain and Hasson, Yana and Lenc, Karel and Mensch, Arthur and Millican, Katherine and Reynolds, Malcolm and others},
  journal={Advances in Neural Information Processing Systems},
  volume={35},
  pages={23716--23736},
  year={2022}
}

@inproceedings{alvar2025divprune,
  title={{DivPrune}: Diversity-Based Visual Token Pruning for Large Multimodal Models},
  author={Alvar, Saeed Ranjbar and Singh, Gursimran and Akbari, Mohammad and Zhang, Yong},
  booktitle={2025 IEEE/CVF Conference on Computer Vision and Pattern Recognition (CVPR)},
  pages={9392--9401},
  year={2025},
  month={June},
  publisher={IEEE},
  doi={10.1109/CVPR52734.2025.00877}
}

@misc{bai2025qwen3vltechnicalreport,
  title={{Qwen3-VL} Technical Report},
  author={Bai, Shuai and Cai, Yuxuan and Chen, Ruizhe and Chen, Keqin and Chen, Xionghui and Cheng, Zesen and Deng, Lianghao and Ding, Wei and Gao, Chang and Ge, Chunjiang and Ge, Wenbin and Guo, Zhifang and Huang, Qidong and Huang, Jie and Huang, Fei and Hui, Binyuan and Jiang, Shutong and Li, Zhaohai and Li, Mingsheng and Li, Mei and Li, Kaixin and Lin, Zicheng and Lin, Junyang and Liu, Xuejing and Liu, Jiawei and Liu, Chenglong and Liu, Yang and Liu, Dayiheng and Liu, Shixuan and Lu, Dunjie and Luo, Ruilin and Lv, Chenxu and Men, Rui and Meng, Lingchen and Ren, Xuancheng and Ren, Xingzhang and Song, Sibo and Sun, Yuchong and Tang, Jun and Tu, Jianhong and Wan, Jianqiang and Wang, Peng and Wang, Pengfei and Wang, Qiuyue and Wang, Yuxuan and Xie, Tianbao and Xu, Yiheng and Xu, Haiyang and Xu, Jin and Yang, Zhibo and Yang, Mingkun and Yang, Jianxin and Yang, An and Yu, Bowen and Zhang, Fei and Zhang, Hang and Zhang, Xi and Zheng, Bo and Zhong, Humen and Zhou, Jingren and Zhou, Fan and Zhou, Jing and Zhu, Yuanzhi and Zhu, Ke},
  year={2025},
  note={arXiv preprint arXiv:2511.21631},
  eprint={2511.21631},
  archivePrefix={arXiv},
  primaryClass={cs.CV},
  url={https://arxiv.org/abs/2511.21631}
}

@misc{yang2024qwen25technicalreport,
  title={{Qwen2.5} Technical Report},
  author={Yang, An and Yang, Baosong and Zhang, Beichen and Hui, Binyuan and Zheng, Bo and Yu, Bowen and Li, Chengyuan and Liu, Dayiheng and Huang, Fei and Wei, Haoran and Lin, Huan and Yang, Jian and Tu, Jianhong and Zhang, Jianwei and Yang, Jianxin and Yang, Jiaxi and Zhou, Jingren and Lin, Junyang and Dang, Kai and Lu, Keming and Bao, Keqin and Yang, Kexin and Yu, Le and Li, Mei and Xue, Mingfeng and Zhang, Pei and Zhu, Qin and Men, Rui and Lin, Runji and Li, Tianhao and Tang, Tianyi and Xia, Tingyu and Ren, Xingzhang and Ren, Xuancheng and Fan, Yang and Su, Yang and Zhang, Yichang and Wan, Yu and Liu, Yuqiong and Cui, Zeyu and Zhang, Zhenru and Qiu, Zihan},
  year={2024},
  note={arXiv preprint arXiv:2412.15115},
  eprint={2412.15115},
  archivePrefix={arXiv},
  primaryClass={cs.CL},
  doi={10.48550/arXiv.2412.15115},
  url={https://arxiv.org/abs/2412.15115}
}

@inproceedings{bolya2022token,
  title={Token Merging: Your {ViT} but Faster},
  author={Bolya, Daniel and Fu, Cheng-Yang and Dai, Xiaoliang and Zhang, Peizhao and Feichtenhofer, Christoph and Hoffman, Judy},
  booktitle={International Conference on Learning Representations},
  year={2023},
  url={https://openreview.net/forum?id=JroZRaRw7Eu}
}

@inproceedings{dos2020good,
  title={The Good, the Bad, and the Ugly: Neural Networks Straight from {JPEG}},
  author={dos Santos, Samuel Felipe and Sebe, Nicu and Almeida, Jurandy},
  booktitle={2020 IEEE International Conference on Image Processing (ICIP)},
  pages={1896--1900},
  year={2020},
  publisher={IEEE},
  doi={10.1109/ICIP40778.2020.9190741},
  url={https://doi.org/10.1109/ICIP40778.2020.9190741}
}

@article{feng2024docpedia,
  title={{DocPedia}: Unleashing the Power of Large Multimodal Model in the Frequency Domain for Versatile Document Understanding},
  author={Feng, Hao and Liu, Qi and Liu, Hao and Tang, Jingqun and Zhou, Wengang and Li, Houqiang and Huang, Can},
  journal={Science China Information Sciences},
  volume={67},
  number={12},
  pages={220106},
  year={2024},
  publisher={Springer},
  doi={10.1007/s11432-024-4250-y},
  url={https://doi.org/10.1007/s11432-024-4250-y}
}

@misc{fu2023mme,
  title={{MME}: A Comprehensive Evaluation Benchmark for Multimodal Large Language Models},
  author={Fu, Chaoyou and Chen, Peixian and Shen, Yunhang and Qin, Yulei and Zhang, Mengdan and Lin, Xu and Yang, Jinrui and Zheng, Xiawu and Li, Ke and Sun, Xing and Wu, Yunsheng and Ji, Rongrong and Shan, Caifeng and He, Ran},
  year={2023},
  note={arXiv preprint arXiv:2306.13394; NeurIPS Datasets and Benchmarks 2025 spotlight},
  eprint={2306.13394},
  archivePrefix={arXiv},
  primaryClass={cs.CV},
  doi={10.48550/arXiv.2306.13394},
  url={https://arxiv.org/abs/2306.13394}
}

@article{hu2024matryoshka,
  title={Matryoshka Query Transformer for Large Vision-Language Models},
  author={Hu, Wenbo and Dou, Zi-Yi and Li, Liunian and Kamath, Amita and Peng, Nanyun and Chang, Kai-Wei},
  journal={Advances in Neural Information Processing Systems},
  volume={37},
  pages={50168--50188},
  year={2024}
}

@inproceedings{hudson2019gqa,
  title={{GQA}: A New Dataset for Real-World Visual Reasoning and Compositional Question Answering},
  author={Hudson, Drew A. and Manning, Christopher D.},
  booktitle={Proceedings of the IEEE/CVF Conference on Computer Vision and Pattern Recognition (CVPR)},
  pages={6700--6709},
  year={2019},
  doi={10.1109/CVPR.2019.00686},
  url={https://openaccess.thecvf.com/content_CVPR_2019/html/Hudson_GQA_A_New_Dataset_for_Real-World_Visual_Reasoning_and_Compositional_CVPR_2019_paper.html}
}

@inproceedings{jaegle2021perceiver,
  title={Perceiver: General Perception with Iterative Attention},
  author={Jaegle, Andrew and Gimeno, Felix and Brock, Andy and Vinyals, Oriol and Zisserman, Andrew and Carreira, Joao},
  booktitle={International Conference on Machine Learning},
  pages={4651--4664},
  year={2021},
  publisher={PMLR}
}

@inproceedings{jain2019attention,
  title={Attention Is Not Explanation},
  author={Jain, Sarthak and Wallace, Byron C.},
  booktitle={Proceedings of the 2019 Conference of the North American Chapter of the Association for Computational Linguistics: Human Language Technologies, Volume 1 (Long and Short Papers)},
  pages={3543--3556},
  year={2019},
  address={Minneapolis, Minnesota},
  publisher={Association for Computational Linguistics},
  doi={10.18653/v1/N19-1357}
}

@inproceedings{li2023blip,
  title={{BLIP-2}: Bootstrapping Language-Image Pre-Training with Frozen Image Encoders and Large Language Models},
  author={Li, Junnan and Li, Dongxu and Savarese, Silvio and Hoi, Steven},
  booktitle={International Conference on Machine Learning},
  pages={19730--19742},
  year={2023},
  publisher={PMLR}
}

@inproceedings{li2023pope,
  title={Evaluating Object Hallucination in Large Vision-Language Models},
  author={Li, Yifan and Du, Yifan and Zhou, Kun and Wang, Jinpeng and Zhao, Xin and Wen, Ji-Rong},
  booktitle={Proceedings of the 2023 Conference on Empirical Methods in Natural Language Processing},
  pages={292--305},
  year={2023},
  month={December},
  address={Singapore},
  publisher={Association for Computational Linguistics},
  doi={10.18653/v1/2023.emnlp-main.20},
  url={https://aclanthology.org/2023.emnlp-main.20/}
}

@inproceedings{li2024inference,
  title={Inference Optimal {VLMs} Need Fewer Visual Tokens and More Parameters},
  author={Li, Kevin Y. and Goyal, Sachin and Semedo, Joao D. and Kolter, J. Zico},
  booktitle={International Conference on Learning Representations},
  year={2025},
  url={https://openreview.net/forum?id=6VhDQP7WGX}
}

@misc{li2024llavaonevision,
  title={{LLaVA-OneVision}: Easy Visual Task Transfer},
  author={Li, Bo and Zhang, Yuanhan and Guo, Dong and Zhang, Renrui and Li, Feng and Zhang, Hao and Zhang, Kaichen and Zhang, Peiyuan and Li, Yanwei and Liu, Ziwei and Li, Chunyuan},
  year={2024},
  note={arXiv preprint arXiv:2408.03326},
  eprint={2408.03326},
  archivePrefix={arXiv},
  primaryClass={cs.CV},
  doi={10.48550/arXiv.2408.03326},
  url={https://arxiv.org/abs/2408.03326}
}

@article{li2025tokenpacker,
  title={{TokenPacker}: Efficient Visual Projector for Multimodal {LLM}},
  author={Li, Wentong and Yuan, Yuqian and Liu, Jian and Tang, Dongqi and Wang, Song and Qin, Jie and Zhu, Jianke and Zhang, Lei},
  journal={International Journal of Computer Vision},
  volume={133},
  number={10},
  pages={6794--6812},
  year={2025},
  month={June},
  publisher={Springer},
  doi={10.1007/s11263-025-02491-7}
}

@inproceedings{liu2015faceattributes,
  title={Deep Learning Face Attributes in the Wild},
  author={Liu, Ziwei and Luo, Ping and Wang, Xiaogang and Tang, Xiaoou},
  booktitle={Proceedings of the IEEE International Conference on Computer Vision (ICCV)},
  pages={3730--3738},
  year={2015},
  doi={10.1109/ICCV.2015.425},
  url={https://openaccess.thecvf.com/content_iccv_2015/html/Liu_Deep_Learning_Face_ICCV_2015_paper.html}
}

@inproceedings{liu2024improved,
  title={Improved Baselines with Visual Instruction Tuning},
  author={Liu, Haotian and Li, Chunyuan and Li, Yuheng and Lee, Yong Jae},
  booktitle={Proceedings of the IEEE/CVF Conference on Computer Vision and Pattern Recognition (CVPR)},
  pages={26296--26306},
  year={2024},
  url={https://openaccess.thecvf.com/content/CVPR2024/html/Liu_Improved_Baselines_with_Visual_Instruction_Tuning_CVPR_2024_paper.html}
}

@inproceedings{liu2024mmbench,
  title={{MMBench}: Is Your Multi-modal Model an All-Around Player?},
  author={Liu, Yuan and Duan, Haodong and Zhang, Yuanhan and Li, Bo and Zhang, Songyang and Zhao, Wangbo and Yuan, Yike and Wang, Jiaqi and He, Conghui and Liu, Ziwei and Chen, Kai and Lin, Dahua},
  booktitle={Computer Vision -- ECCV 2024},
  series={Lecture Notes in Computer Science},
  volume={15064},
  pages={216--233},
  year={2024},
  publisher={Springer},
  doi={10.1007/978-3-031-72658-3_13},
  url={https://doi.org/10.1007/978-3-031-72658-3_13}
}

@inproceedings{liang2022not,
  title={{EViT}: Expediting Vision Transformers via Token Reorganizations},
  author={Liang, Youwei and Ge, Chongjian and Tong, Zhan and Song, Yibing and Wang, Jue and Xie, Pengtao},
  booktitle={International Conference on Learning Representations},
  year={2022},
  url={https://openreview.net/forum?id=BjyvwnXXVn_}
}

@inproceedings{lu2022scienceqa,
  title={Learn to Explain: Multimodal Reasoning via Thought Chains for Science Question Answering},
  author={Lu, Pan and Mishra, Swaroop and Xia, Tanglin and Qiu, Liang and Chang, Kai-Wei and Zhu, Song-Chun and Tafjord, Oyvind and Clark, Peter and Kalyan, Ashwin},
  booktitle={Advances in Neural Information Processing Systems},
  volume={35},
  pages={2507--2521},
  year={2022},
  url={https://proceedings.neurips.cc/paper_files/paper/2022/hash/11332b6b6cf4485b84afadb1352d3a9a-Abstract-Conference.html}
}

@article{mallat1989multiresolution,
  title={Multiresolution Approximations and Wavelet Orthonormal Bases of {$L^2(\mathbb{R})$}},
  author={Mallat, Stephane G.},
  journal={Transactions of the American Mathematical Society},
  volume={315},
  number={1},
  pages={69--87},
  year={1989},
  doi={10.1090/S0002-9947-1989-1008470-5},
  url={https://doi.org/10.1090/S0002-9947-1989-1008470-5}
}

@inproceedings{qin2021fcanet,
  title={{FcaNet}: Frequency Channel Attention Networks},
  author={Qin, Zequn and Zhang, Pengyi and Wu, Fei and Li, Xi},
  booktitle={Proceedings of the IEEE/CVF International Conference on Computer Vision},
  pages={763--772},
  year={2021},
  doi={10.1109/ICCV48922.2021.00082},
  url={https://openaccess.thecvf.com/content/ICCV2021/html/Qin_FcaNet_Frequency_Channel_Attention_Networks_ICCV_2021_paper.html}
}

@article{rao2021dynamicvit,
  title={{DynamicViT}: Efficient Vision Transformers with Dynamic Token Sparsification},
  author={Rao, Yongming and Zhao, Wenliang and Liu, Benlin and Lu, Jiwen and Zhou, Jie and Hsieh, Cho-Jui},
  journal={Advances in Neural Information Processing Systems},
  volume={34},
  pages={13937--13949},
  year={2021}
}

@article{ryoo2021tokenlearner,
  title={{TokenLearner}: Adaptive Space-Time Tokenization for Videos},
  author={Ryoo, Michael and Piergiovanni, A. J. and Arnab, Anurag and Dehghani, Mostafa and Angelova, Anelia},
  journal={Advances in Neural Information Processing Systems},
  volume={34},
  pages={12786--12797},
  year={2021}
}

@article{salimans2016weight,
  title={Weight Normalization: A Simple Reparameterization to Accelerate Training of Deep Neural Networks},
  author={Salimans, Tim and Kingma, Durk P.},
  journal={Advances in Neural Information Processing Systems},
  volume={29},
  year={2016}
}

@inproceedings{shang_llava-prumerge_2024,
  title={{LLaVA-PruMerge}: Adaptive Token Reduction for Efficient Large Multimodal Models},
  author={Shang, Yuzhang and Cai, Mu and Xu, Bingxin and Lee, Yong Jae and Yan, Yan},
  booktitle={Proceedings of the IEEE/CVF International Conference on Computer Vision},
  pages={22857--22867},
  year={2025},
  month={October}
}

@inproceedings{singh2019textvqa,
  title={Towards {VQA} Models That Can Read},
  author={Singh, Amanpreet and Natarajan, Vivek and Shah, Meet and Jiang, Yu and Chen, Xinlei and Batra, Dhruv and Parikh, Devi and Rohrbach, Marcus},
  booktitle={Proceedings of the IEEE/CVF Conference on Computer Vision and Pattern Recognition (CVPR)},
  pages={8317--8326},
  year={2019},
  doi={10.1109/CVPR.2019.00851},
  url={https://openaccess.thecvf.com/content_CVPR_2019/html/Singh_Towards_VQA_Models_That_Can_Read_CVPR_2019_paper.html}
}

@article{stewart1980efficient,
  title={The Efficient Generation of Random Orthogonal Matrices with an Application to Condition Estimators},
  author={Stewart, Gilbert W.},
  journal={SIAM Journal on Numerical Analysis},
  volume={17},
  number={3},
  pages={403--409},
  year={1980},
  publisher={SIAM},
  doi={10.1137/0717034},
  url={https://doi.org/10.1137/0717034}
}

@inproceedings{martins2016sparsemax,
  title={From Softmax to Sparsemax: A Sparse Model of Attention and Multi-Label Classification},
  author={Martins, Andre and Astudillo, Ramon},
  booktitle={Proceedings of the 33rd International Conference on Machine Learning},
  pages={1614--1623},
  year={2016},
  volume={48},
  series={Proceedings of Machine Learning Research},
  address={New York, New York, USA},
  month={20--22 Jun},
  publisher={PMLR},
  url={https://proceedings.mlr.press/v48/martins16.html}
}

@misc{vteam2026glm45vglm41vthinkingversatilemultimodal,
  title={{GLM-4.5V} and {GLM-4.1V-Thinking}: Towards Versatile Multimodal Reasoning with Scalable Reinforcement Learning},
  author={{GLM-V Team} and Hong, Wenyi and Yu, Wenmeng and Gu, Xiaotao and Wang, Guo and Gan, Guobing and Tang, Haomiao and Cheng, Jiale and Qi, Ji and Ji, Junhui and Pan, Lihang and Duan, Shuaiqi and Wang, Weihan and Wang, Yan and Cheng, Yean and He, Zehai and Su, Zhe and Yang, Zhen and Pan, Ziyang and Zeng, Aohan and Wang, Baoxu and Chen, Bin and Shi, Boyan and Pang, Changyu and Zhang, Chenhui and Yin, Da and Yang, Fan and Chen, Guoqing and Li, Haochen and Zhu, Jiale and Chen, Jiali and Xu, Jiaxing and Xu, Jiazheng and Chen, Jing and Lin, Jinghao and Chen, Jinhao and Wang, Jinjiang and Chen, Junjie and Lei, Leqi and Gong, Letian and Pan, Leyi and Liu, Mingdao and Xu, Mingde and Zhang, Mingzhi and Zheng, Qinkai and Lyu, Ruiliang and Tu, Shangqin and Yang, Sheng and Meng, Shengbiao and Zhong, Shi and Huang, Shiyu and Zhao, Shuyuan and Xue, Siyan and Zhang, Tianshu and Luo, Tianwei and Hao, Tianxiang and Tong, Tianyu and Jia, Wei and Li, Wenkai and Liu, Xiao and Zhang, Xiaohan and Lyu, Xin and Zhang, Xinyu and Fan, Xinyue and Huang, Xuancheng and Xue, Yadong and Wang, Yanfeng and Wang, Yanling and Wang, Yanzi and An, Yifan and Du, Yifan and Huang, Yiheng and Niu, Yilin and Shi, Yiming and Wang, Yu and Wang, Yuan and Yue, Yuanchang and Li, Yuchen and Liu, Yusen and Zhang, Yutao and Wang, Yuting and Zhang, Yuxuan and Xue, Zhao and Du, Zhengxiao and Hou, Zhenyu and Wang, Zihan and Zhang, Peng and Liu, Debing and Xu, Bin and Li, Juanzi and Huang, Minlie and Dong, Yuxiao and Tang, Jie},
  year={2025},
  note={arXiv preprint arXiv:2507.01006; version 6 updated on 2026-01-01},
  eprint={2507.01006},
  archivePrefix={arXiv},
  primaryClass={cs.CV},
  doi={10.48550/arXiv.2507.01006},
  url={https://arxiv.org/abs/2507.01006}
}

@misc{vicuna2023,
  title={Vicuna: An Open-Source Chatbot Impressing {GPT-4} with 90\% {ChatGPT} Quality},
  author={Chiang, Wei-Lin and Li, Zhuohan and Lin, Zi and Sheng, Ying and Wu, Zhanghao and Zhang, Hao and Zheng, Lianmin and Zhuang, Siyuan and Zhuang, Yonghao and Gonzalez, Joseph E. and Stoica, Ion and Xing, Eric P.},
  year={2023},
  month={March},
  howpublished={Blog post},
  url={https://lmsys.org/blog/2023-03-30-vicuna/}
}

@article{wallace1991jpeg,
  title={The {JPEG} Still Picture Compression Standard},
  author={Wallace, Gregory K.},
  journal={Communications of the ACM},
  volume={34},
  number={4},
  pages={30--44},
  year={1991},
  publisher={ACM},
  doi={10.1145/103085.103089},
  url={https://doi.org/10.1145/103085.103089}
}

@misc{wang2025fourier,
  title={{Fourier-VLM}: Compressing Vision Tokens in the Frequency Domain for Large Vision-Language Models},
  author={Wang, Huanyu and Kai, Jushi and Bai, Haoli and Hou, Lu and Jiang, Bo and He, Ziwei and Lin, Zhouhan},
  year={2025},
  note={arXiv preprint arXiv:2508.06038},
  eprint={2508.06038},
  archivePrefix={arXiv},
  primaryClass={cs.CV},
  url={https://arxiv.org/abs/2508.06038}
}

@inproceedings{wen2025stop,
  title={Stop Looking for Important Tokens in Multimodal Language Models: Duplication Matters More},
  author={Wen, Zichen and Gao, Yifeng and Wang, Shaobo and Zhang, Junyuan and Zhang, Qintong and Li, Weijia and He, Conghui and Zhang, Linfeng},
  booktitle={Proceedings of the 2025 Conference on Empirical Methods in Natural Language Processing},
  pages={9961--9980},
  year={2025},
  month={November},
  address={Suzhou, China},
  publisher={Association for Computational Linguistics},
  doi={10.18653/v1/2025.emnlp-main.505},
  url={https://aclanthology.org/2025.emnlp-main.505/}
}

@misc{wu2024deepseek,
  title={{DeepSeek-VL2}: Mixture-of-Experts Vision-Language Models for Advanced Multimodal Understanding},
  author={Wu, Zhiyu and Chen, Xiaokang and Pan, Zizheng and Liu, Xingchao and Liu, Wen and Dai, Damai and Gao, Huazuo and Ma, Yiyang and Wu, Chengyue and Wang, Bingxuan and others},
  year={2024},
  note={arXiv preprint arXiv:2412.10302},
  eprint={2412.10302},
  archivePrefix={arXiv},
  primaryClass={cs.CV},
  url={https://arxiv.org/abs/2412.10302}
}

@inproceedings{yu2024mmvet,
  title={{MM}-Vet: Evaluating Large Multimodal Models for Integrated Capabilities},
  author={Yu, Weihao and Yang, Zhengyuan and Li, Linjie and Wang, Jianfeng and Lin, Kevin and Liu, Zicheng and Wang, Xinchao and Wang, Lijuan},
  booktitle={Proceedings of the 41st International Conference on Machine Learning},
  series={Proceedings of Machine Learning Research},
  volume={235},
  pages={57730--57754},
  year={2024},
  publisher={PMLR},
  url={https://proceedings.mlr.press/v235/yu24o.html}
}

@inproceedings{zhang_beyond_2025-1,
  title={Beyond Attention or Similarity: Maximizing Conditional Diversity for Token Pruning in {MLLMs}},
  author={Zhang, Qizhe and Liu, Mengzhen and Li, Lichen and Lu, Ming and Zhang, Yuan and Pan, Junwen and She, Qi and Zhang, Shanghang},
  booktitle={Advances in Neural Information Processing Systems},
  year={2025},
  url={https://openreview.net/forum?id=BLLixcuZgl}
}

@inproceedings{zhang2025llava,
  title={{LLaVA-Mini}: Efficient Image and Video Large Multimodal Models with One Vision Token},
  author={Zhang, Shaolei and Fang, Qingkai and Yang, Zhe and Feng, Yang},
  booktitle={International Conference on Learning Representations},
  year={2025},
  url={https://openreview.net/forum?id=UQJ7CDW8nb}
}

@misc{zhu2025internvl3,
  title={{InternVL3}: Exploring Advanced Training and Test-Time Recipes for Open-Source Multimodal Models},
  author={Zhu, Jinguo and Wang, Weiyun and Chen, Zhe and Liu, Zhaoyang and Ye, Shenglong and Gu, Lixin and Tian, Hao and Duan, Yuchen and Su, Weijie and Shao, Jie and others},
  year={2025},
  note={arXiv preprint arXiv:2504.10479},
  eprint={2504.10479},
  archivePrefix={arXiv},
  primaryClass={cs.CV},
  url={https://arxiv.org/abs/2504.10479}
}

@misc{tschannen2025siglip2,
  title={{SigLIP 2}: Multilingual Vision-Language Encoders with Improved Semantic Understanding, Localization, and Dense Features},
  author={Tschannen, Michael and Gritsenko, Alexey and Wang, Xiao and Naeem, Muhammad Ferjad and Alabdulmohsin, Ibrahim and Parthasarathy, Nikhil and Evans, Talfan and Beyer, Lucas and Xia, Ye and Mustafa, Basil and H{\'e}naff, Olivier and Harmsen, Jeremiah and Steiner, Andreas and Zhai, Xiaohua},
  year={2025},
  note={arXiv preprint arXiv:2502.14786},
  eprint={2502.14786},
  archivePrefix={arXiv},
  url={https://arxiv.org/abs/2502.14786}
}

@misc{zhai2023siglip,
  title={Sigmoid Loss for Language Image Pre-Training},
  author={Zhai, Xiaohua and Mustafa, Basil and Kolesnikov, Alexander and Beyer, Lucas},
  year={2023},
  note={arXiv preprint arXiv:2303.15343},
  eprint={2303.15343},
  archivePrefix={arXiv},
  url={https://arxiv.org/abs/2303.15343}
}

@inproceedings{mathew2021docvqa,
  title={{DocVQA}: A Dataset for {VQA} on Document Images},
  author={Mathew, Minesh and Karatzas, Dimosthenis and Jawahar, C. V.},
  booktitle={Proceedings of the IEEE/CVF Winter Conference on Applications of Computer Vision (WACV)},
  year={2021},
  pages={2200--2209}
}
}

\newpage
\appendix
\section{Derivation of the energy-retention term in the compressibility functional}
\label{app:energy_propto}

This theory appendix follows the methodology in \cref{sec:method}.  It first derives the compressibility terms, then the
learnability objective, and finally the concrete Braco design choices: the structured low-pass set, basis baselines,
basis-coordinate embedding, coordinate organization, and spatial residual domain and implementation details.

The main text reports the normalized-trace energy score in \cref{eq:unified_energy_retention}.  This section derives that
form from the token-axis second moment
\(\mathcal{E}_o(\mathbf{B};\mathcal{S};\bar{\mathbf{Z}}_i)=\bar{\mathbf{Z}}_i\bar{\mathbf{Z}}_i^\top\) by scalarizing the
matrix second moment with \(\mathrm{tr}(\cdot)\) and normalizing by the total expected token energy.

\paragraph{Retained energy.}
Recall \(\bar{\mathbf{Z}}_i=\mathbf{P}_{\mathcal{S}}\mathbf{U}_{\mathbf{B}}\mathbf{X}_i\in\mathbb{R}^{K\times D_v}\).
Here \(\mathbf{X}_i\) is the \(i\)-th visual-token grid, \(\mathbf{U}_{\mathbf{B}}\) is the token-lattice transform induced
by basis \(\mathbf{B}\), \(\mathbf{P}_{\mathcal{S}}\) selects the retained index set \(\mathcal{S}\), \(K=|\mathcal{S}|\),
and \(D_v\) is the token dimension.
A canonical scalar notion of ``retained energy'' is the squared Frobenius norm
\begin{equation}
\|\bar{\mathbf{Z}}_i\|_F^2
\;=\;
\mathrm{tr}(\bar{\mathbf{Z}}_i\bar{\mathbf{Z}}_i^\top)
\;=\;
\mathrm{tr}\!\Big(
\mathbf{P}_{\mathcal{S}}\mathbf{U}_{\mathbf{B}}\mathbf{X}_i\mathbf{X}_i^\top
\mathbf{U}_{\mathbf{B}}^\top\mathbf{P}_{\mathcal{S}}^\top
\Big),
\label{eq:app_retained_energy_single}
\end{equation}
where we used the cyclic trace identity \(\mathrm{tr}(\mathbf{A}\mathbf{B})=\mathrm{tr}(\mathbf{B}\mathbf{A})\) for
conformable matrices.
Taking expectation over \(\mathbf{X}_i\sim\mathcal{D}\) and defining
\(\mathbf{M}\triangleq \mathbb{E}[\mathbf{X}_i\mathbf{X}_i^\top]\in\mathbb{R}^{L\times L}\) as in \cref{eq:unified_energy_retention}, we obtain
\begin{equation}
\mathbb{E}\|\bar{\mathbf{Z}}_i\|_F^2
\;=\;
\mathrm{tr}\!\Big(
\mathbf{P}_{\mathcal{S}}\mathbf{U}_{\mathbf{B}}\mathbf{M}
\mathbf{U}_{\mathbf{B}}^\top\mathbf{P}_{\mathcal{S}}^\top
\Big)
\;=\;
\mathrm{tr}\!\left(\mathbb{E}[\bar{\mathbf{Z}}_i\bar{\mathbf{Z}}_i^\top]\right).
\label{eq:app_retained_energy_expect}
\end{equation}
Thus, the trace of the matrix second moment is exactly the expected retained energy under the fixed index set
\(\mathcal{S}\).

\paragraph{Normalization by total expected energy.}
Similarly, the expected total energy of the original token grid is
\begin{equation}
\mathbb{E}\|\mathbf{X}_i\|_F^2
\;=\;
\mathbb{E}\,\mathrm{tr}(\mathbf{X}_i\mathbf{X}_i^\top)
\;=\;
\mathrm{tr}(\mathbf{M}).
\label{eq:app_total_energy_expect}
\end{equation}
Therefore, the \emph{fraction} of energy preserved by the basis--truncation pair \((\mathbf{B},\mathcal{S})\) admits the
normalized form
\begin{equation}
\frac{\mathbb{E}\|\bar{\mathbf{Z}}_i\|_F^2}{\mathbb{E}\|\mathbf{X}_i\|_F^2}
\;=\;
\frac{
\mathrm{tr}\!\Big(
\mathbf{P}_{\mathcal{S}}\mathbf{U}_{\mathbf{B}}\mathbf{M}
\mathbf{U}_{\mathbf{B}}^\top\mathbf{P}_{\mathcal{S}}^\top
\Big)}
{\mathrm{tr}(\mathbf{M})},
\label{eq:app_energy_ratio}
\end{equation}
which is exactly the scalar quantity used in \cref{eq:unified_energy_retention}.  It is a dimensionless score and is
comparable across choices of \((\mathbf{B},\mathcal{S})\) for the same token-field distribution.

\paragraph{Interpretation as energy restricted to a fixed retained subspace.}
Let \(\mathbf{S}_{\mathcal{S}}\triangleq \mathbf{P}_{\mathcal{S}}^\top\mathbf{P}_{\mathcal{S}}\in\mathbb{R}^{L\times L}\),
which is the coordinate projector onto the fixed index set \(\mathcal{S}\) in the transformed domain.
Then the numerator in \cref{eq:app_energy_ratio} can be rewritten as
\(
\mathrm{tr}\big(\mathbf{U}_{\mathbf{B}}\mathbf{M}\mathbf{U}_{\mathbf{B}}^\top\mathbf{S}_{\mathcal{S}}\big)
\),
i.e., the expected energy of \(\mathbf{U}_{\mathbf{B}}\mathbf{X}_i\) \emph{restricted} to the fixed retained coordinates
\(\mathcal{S}\).
Since \(\mathbf{U}_{\mathbf{B}}\) is orthonormal (because \(\mathbf{B}\) is orthonormal and \(\mathbf{U}_{\mathbf{B}}
=\mathbf{B}\otimes\mathbf{B}\)), the transform itself preserves total energy:
\(\mathbb{E}\|\mathbf{U}_{\mathbf{B}}\mathbf{X}_i\|_F^2=\mathbb{E}\|\mathbf{X}_i\|_F^2=\mathrm{tr}(\mathbf{M})\);
the only energy loss comes from the structured truncation \(\mathbf{P}_{\mathcal{S}}\).

\paragraph{Invariance to orthogonal coordinate organization \(\mathbf{A}\).}
Finally, if we further apply an orthogonal re-parameterization inside the retained subspace
\(\mathbf{Z}_i=\mathbf{A}\bar{\mathbf{Z}}_i\) with \(\mathbf{A}^\top\mathbf{A}=\mathbf{I}_K\), then
\begin{equation}
\|\mathbf{Z}_i\|_F^2
=
\mathrm{tr}(\mathbf{A}\bar{\mathbf{Z}}_i\bar{\mathbf{Z}}_i^\top\mathbf{A}^\top)
=
\mathrm{tr}(\bar{\mathbf{Z}}_i\bar{\mathbf{Z}}_i^\top\mathbf{A}^\top\mathbf{A})
=
\|\bar{\mathbf{Z}}_i\|_F^2,
\label{eq:app_A_invariance}
\end{equation}
so the energy-retention score in \cref{eq:unified_energy_retention} depends only on \((\mathbf{B},\mathcal{S})\) (the retained \emph{subspace}),
and is unaffected by \(\mathbf{A}\) (which only changes the \emph{coordinates} inside that subspace).

\section{Task-readability term in the compressibility functional}
\label{app:readability_term}

Energy retention is a useful distortion proxy, but it does not by itself say whether the retained subspace contains
task-relevant evidence.  To formalize this complementary notion, model the downstream task locally by a linear target
\begin{equation}
f^\star(\mathbf{X})=\langle\mathbf{W}^\star,\mathbf{X}\rangle,\qquad
\langle\mathbf{W},\mathbf{X}\rangle=\mathrm{tr}(\mathbf{W}^\top\mathbf{X}),
\end{equation}
where \(\mathbf{W}^\star\in\mathbb{R}^{L\times D_v}\) is the task direction.  The set of original token fields
representable using only the retained transform coordinates is
\begin{equation}
\mathcal{U}(\mathbf{B},\mathcal{S})
\triangleq
\left\{
\mathbf{U}_{\mathbf{B}}^\top\mathbf{P}_{\mathcal{S}}^\top\mathbf{Q}
:\mathbf{Q}\in\mathbb{R}^{K\times D_v}
\right\}.
\label{eq:unified_subspace}
\end{equation}
Let \(\mathbf{\Pi}_{\mathbf{B},\mathcal{S}}\) be the \(\mathbf{M}\)-orthogonal projector onto
\(\mathcal{U}(\mathbf{B},\mathcal{S})\), and let \(\|\cdot\|_{\mathbf{M}}\) denote the induced \(\mathbf{M}\)-weighted
norm.  We define readability as the fraction of task-direction power retained by the subspace:
\begin{equation}
\mathcal{R}(\mathbf{B};\mathcal{S})
\triangleq
\frac{
\mathbb{E}\left\|
\mathbf{\Pi}_{\mathbf{B},\mathcal{S}}\mathbf{W}^\star
\right\|_{\mathbf{M}}^{2}}
{
\mathbb{E}\left\|
\mathbf{W}^\star
\right\|_{\mathbf{M}}^{2}}.
\label{eq:unified_R}
\end{equation}
By construction, \(\mathcal{R}\in[0,1]\), and larger values indicate that the retained subspace is better aligned with
the task direction.  The main-text compressibility score \(\mathcal{C}\) combines this readability term with energy
retention.

\section{Statistical conditioning term}
\label{app:stat_cond_term}
We use a rotation-sensitive surrogate that favors weak cross-token correlations and balanced token scales.
For retained coordinates $\bar{\mathbf{Z}}$, let
$\bar{\mathbf{G}}=\mathbb{E}[\bar{\mathbf{Z}}\bar{\mathbf{Z}}^\top]$ be the pre-organization token Gram matrix and
$\mathbf{G}(\mathbf{A})=\mathbf{A}\bar{\mathbf{G}}\mathbf{A}^\top$ be the Gram matrix after applying the coordinate
organization $\mathbf{A}$.
Since $\mathbf{A}$ is orthogonal, $\mathrm{tr}(\mathbf{G}(\mathbf{A}))=\mathrm{tr}(\bar{\mathbf{G}})$; rotations can
therefore change token correlations and scale balance while preserving total token variance.
\begin{equation}
\begin{aligned}
\mathcal{L}_{\mathrm{st}}(\mathbf{A})
\;\triangleq\;&\;
\big\|\mathrm{OffDiag}(\mathbf{G}(\mathbf{A}))\big\|_F^2 \\
&\;+\;
\gamma\,
\Big\|
\mathrm{diag}(\mathbf{G}(\mathbf{A}))
-
\frac{\mathrm{tr}(\bar{\mathbf{G}})}{K}\,\mathbf{1}
\Big\|_2^2,
\end{aligned}
\label{eq:app_unified_stat_obj}
\end{equation}
where $\gamma>0$ is a fixed weight, $\mathbf{1}\in\mathbb{R}^{K}$ is all-ones, $\mathrm{OffDiag}(\cdot)$ zeroes diagonal
entries, and $\mathrm{diag}(\cdot)$ extracts diagonal entries.
The first term penalizes off-diagonal correlations; the second penalizes deviations of per-token variance from the
uniform average $\mathrm{tr}(\bar{\mathbf{G}})/K$.

\section{Final learnability objective}
\label{app:final_learn_obj}
To compare learnability across token budgets, we adjust the relative strength of statistical conditioning and geometric alignment.
Since the statistical term \(\mathcal{L}_{\mathrm{st}}\) is computed from a \(K\times K\) Gram matrix, its raw magnitude grows with \(K\); we therefore normalize it by \(K^2\) to keep it comparable across budgets.
The geometric term \(\mathcal{L}_{\mathrm{geo}}\) is already normalized by construction (see \cref{eq:unified_geom_obj}), so we keep it unscaled and tune its relative importance with \(\beta\).
This yields the budget-aware objective:
\begin{equation}
\boxed{
\mathcal{L}_{\mathrm{learn}}(\mathbf{A};K)
\triangleq
\frac{1}{K^2}\,\mathcal{L}_{\mathrm{st}}(\mathbf{A})+
\beta\mathcal{L}_{\mathrm{geo}}(\mathbf{A}).
}
\qquad
\label{eq:app_unified_learn_obj}
\end{equation}
The scaling in \cref{eq:app_unified_learn_obj} keeps the statistical conditioning term comparable across token budgets, while \(\beta\) controls the trade-off with geometric compatibility.

\section{\texorpdfstring{Choosing $\mathcal{S}$: the $C\times C$ low-pass block}{Choosing S: the C x C low-pass block}}
\label{app:choosing_S}
For separable transforms with lattice coordinates $(u,v)\in\{0,\dots,N-1\}^2$, the canonical deployable structured
low-pass rule is the tensor-product block
\begin{equation}
\mathcal{S}_C
\;=\;
\{(u,v): 0\le u<C,\ 0\le v<C\},
\qquad
K_b=C^2,
\label{eq:specialize_block}
\end{equation}
where $C$ is the retained frequency cut-off per axis, and $K_b$ is the backbone token count.
In implementation, $\mathcal{S}_C$ is mapped to a length-$K_b$ index set in $[L]$ via a fixed flattening order of the
$(u,v)$ lattice.

\section{Baselines and qualitative implications}
\label{app:baselines}
\textbf{randortho.} As an isotropic reference, consider a full Haar-random orthogonal transform
$\mathbf{U}\in\mathbb{R}^{L\times L}$ independent of $\mathbf{M}$.  A fixed selection of $K_b$ coordinates retains
the proportional expected energy share:
\begin{equation}
\mathbb{E}_{\mathbf{U}}\!\left[
\frac{\mathrm{tr}(\mathbf{P}_{\mathcal{S}_C}\mathbf{U}\mathbf{M}\mathbf{U}^\top\mathbf{P}_{\mathcal{S}_C}^\top)}
{\mathrm{tr}(\mathbf{M})}\right]
\;=\;
\frac{K_b}{L},
\label{eq:specialize_randC}
\end{equation}
since $\mathbb{E}_{\mathbf{U}}[\mathbf{U}\mathbf{M}\mathbf{U}^\top]
=\mathrm{tr}(\mathbf{M})\mathbf{I}_L/L$.
Our separable \texttt{randortho} baseline uses $\mathbf{Q}_N\otimes\mathbf{Q}_N$ for a Haar-random orthogonal
$\mathbf{Q}_N\in\mathbb{R}^{N\times N}$, with its energy retention and task readability evaluated empirically in
\cref{fig:energy_curves,fig:exp_compressibility_diag}.

\textbf{spatial.} With $\mathbf{B}=\mathbf{I}_N$, where $\mathbf{I}_N$ is the $N\times N$ identity basis,
$\mathcal{S}_C$ is a fixed spatial mask.
Then \eqref{eq:unified_energy_retention} becomes
$\mathcal{E}(\mathbf{I}_N;\mathcal{S}_C)
=\mathrm{tr}(\mathbf{P}_{\mathcal{S}_C}\mathbf{M}\mathbf{P}_{\mathcal{S}_C}^\top)/\mathrm{tr}(\mathbf{M})$,
which is large only if $\mathbf{M}$ concentrates mass on those selected locations; the same limitation carries to
$\mathcal{R}(\mathbf{I}_N;\mathcal{S}_C)$.

\textbf{haar.} With $\mathbf{B}=\mathbf{H}_N$, where $\mathbf{H}_N$ is the $N\times N$ orthonormal Haar basis, Haar
coordinates capture multiscale structure, but the fixed low-pass block $\mathcal{S}_C$ does not
implement tree-structured selections typically favored by Haar, limiting $\mathcal{C}(\mathbf{H}_N;\mathcal{S}_C)$ under
our deployable rule.

\textbf{dct.} With $\mathbf{B}=\mathbf{D}_N$, where $\mathbf{D}_N$ is the $N\times N$ orthonormal cosine basis, token
fields with dominant short-range correlations make $\mathbf{M}$ well-approximated by a structured
operator whose eigenvectors are close to cosine modes.
In this regime, the DCT approximately diagonalizes $\mathbf{M}$ and concentrates energy toward low frequencies, directly
increasing both $\mathcal{E}(\mathbf{D}_N;\mathcal{S}_C)$ and $\mathcal{R}(\mathbf{D}_N;\mathcal{S}_C)$ under the same
$\mathcal{S}_C$.

\section{Basis-coordinate embedding details}
\label{app:basis_coordinate_embedding}

A structured truncation in the basis space removes the explicit spatial coordinates that a downstream multimodal projector
would otherwise receive from the original patch grid.  Braco therefore adds an input-independent embedding on the retained
transform lattice before serialization.

For a lattice coordinate $(u,v)\in\{0,\dots,N-1\}^2$, define
\begin{equation}
r=\frac{\sqrt{u^2+v^2}}{\sqrt{2}\,(N-1)}, \qquad
\theta=\mathrm{atan2}(v,u),
\label{eq:specialize_polar_coords}
\end{equation}
where $r$ is the normalized radial frequency and $\theta$ is the polar angle, with $\theta=0$ at $(u,v)=(0,0)$ by
convention.  With $F$ Fourier frequencies $\omega_k=2^k$, the polar Fourier feature is
\begin{equation}
\boldsymbol{\phi}(u,v)=
\bigl[
\{\sin(\pi\omega_k r),\cos(\pi\omega_k r)\}_{k=0}^{F-1},
\{\sin(\omega_k\theta),\cos(\omega_k\theta)\}_{k=0}^{F-1}
\bigr]^\top .
\label{eq:specialize_polar_pe}
\end{equation}
A shared linear map $\mathbf{W}_{\mathrm{pe}}\in\mathbb{R}^{D_v\times 4F}$ projects
$\boldsymbol{\phi}(u,v)\in\mathbb{R}^{4F}$ into the token feature dimension, and a learnable scalar gate $\alpha$
controls the embedding magnitude.  If $\mathbf{Y}_{u,v,:}$ is the retained transform token at coordinate $(u,v)$, the
embedded token $\tilde{\mathbf{Y}}_{u,v,:}$ is
\begin{equation}
\tilde{\mathbf{Y}}_{u,v,:}
=
\mathbf{Y}_{u,v,:}
+
\alpha\,\mathbf{W}_{\mathrm{pe}}\boldsymbol{\phi}(u,v).
\label{eq:coder_add_pe}
\end{equation}
This embedding is independent of the input image and serves only as a stable coordinate code for the retained transform
tokens.

\section{Coordinate organization candidates}
\label{app:coord_candidates}
In this section, $K_b=C^2$ denotes the retained backbone subspace size before adding spatial residual tokens.
(i) $\mathbf{I}\in\mathbb{R}^{K_b\times K_b}$ keeps the retained DCT coefficients as tokens (\texttt{vanilla});
(ii) $\mathbf{U}_C\in\mathbb{R}^{K_b\times K_b}$ is the orthogonal inverse-DCT map restricted to the retained $C\times C$
subspace, converting coefficients into a $C\times C$ coarse spatial grid which is then serialized (\texttt{idct});
and (iii) $\mathbf{R}\in\mathbb{R}^{K_b\times K_b}$ is a fixed Haar-random orthogonal rotation.
In this setting, the natural preferred organization for downstream geometry is the coarse-grid layout, so we set
\begin{equation}
\mathbf{A}_0 \;=\; \mathbf{U}_C.
\label{eq:specialize_A0}
\end{equation}

\paragraph{Substituting the Gram transform.}
Using $\mathbf{G}(\mathbf{A})=\mathbf{A}\bar{\mathbf{G}}\mathbf{A}^\top$ and \eqref{eq:unified_learn_obj}, we obtain
\begin{equation}
\mathcal{L}_{\mathrm{learn}}(\mathbf{A};K_b)
=
\frac{1}{K_b^2}\,\mathcal{L}_{\mathrm{st}}(\mathbf{A})
+
\beta
\Big(
1-\frac{1}{K_b}\langle \mathbf{A},\mathbf{U}_C\rangle_F
\Big),
\label{eq:specialize_learn_subst_new}
\end{equation}
where $\mathcal{L}_{\mathrm{st}}$ is defined in \eqref{eq:app_unified_stat_obj} and
$\langle \mathbf{A},\mathbf{U}_C\rangle_F=\mathrm{tr}(\mathbf{A}^\top\mathbf{U}_C)$.

\paragraph{\texttt{vanilla} vs.\ \texttt{idct}: an explicit budget threshold.}
Under low-pass DCT truncation, $\bar{\mathbf{G}}$ is often closer to diagonal in coefficient coordinates than after dense
mixings.
Thus $\mathbf{A}=\mathbf{I}$ tends to reduce cross-token correlations (the off-diagonal penalty in
$\mathcal{L}_{\mathrm{st}}$), while $\mathbf{A}=\mathbf{U}_C$ improves geometric compatibility but usually densifies
$\mathbf{G}(\mathbf{A})$.

Define the data-dependent statistical gap and geometric mismatch
\begin{equation}
\begin{aligned}
\Delta_{\mathrm{st}}
\;&\triangleq\;
\mathcal{L}_{\mathrm{st}}(\mathbf{U}_C)
-
\mathcal{L}_{\mathrm{st}}(\mathbf{I}), \\
\rho_C
\;&\triangleq\;
\mathcal{L}_{\mathrm{geo}}(\mathbf{I})
=
1-\frac{1}{K_b}\langle \mathbf{I},\mathbf{U}_C\rangle_F,
\end{aligned}
\label{eq:specialize_deltas}
\end{equation}
where $\Delta_{\mathrm{st}}$ measures how much statistical conditioning worsens when moving from \texttt{vanilla} to
\texttt{idct}, and $\rho_C\in[0,2]$ is the geometric penalty incurred by staying in coefficient coordinates.

Since $\mathcal{L}_{\mathrm{geo}}(\mathbf{U}_C)=0$ by \eqref{eq:unified_geom_obj}, subtracting
$\mathcal{L}_{\mathrm{learn}}(\mathbf{I};K_b)$ from $\mathcal{L}_{\mathrm{learn}}(\mathbf{U}_C;K_b)$ yields
\begin{equation}
\mathcal{L}_{\mathrm{learn}}(\mathbf{U}_C;K_b)
-
\mathcal{L}_{\mathrm{learn}}(\mathbf{I};K_b)
=
\frac{1}{K_b^2}\,\Delta_{\mathrm{st}}
-
\beta\,\rho_C.
\label{eq:specialize_compare}
\end{equation}
When $\Delta_{\mathrm{st}}>0$ and $\rho_C>0$, this two-candidate surrogate comparison admits a crossover threshold:
\begin{equation}
\mathbf{A}^\star=
\begin{cases}
\mathbf{I} & \text{if } K_b^2 < \dfrac{\Delta_{\mathrm{st}}}{\beta\,\rho_C},
\\[2.5mm]
\mathbf{U}_C & \text{if } K_b^2 > \dfrac{\Delta_{\mathrm{st}}}{\beta\,\rho_C}.
\end{cases}
\label{eq:specialize_threshold}
\end{equation}
Equation~\eqref{eq:specialize_threshold} formalizes the intended comparison between \(\mathbf{I}\) and \(\mathbf{U}_C\):
\textbf{when $K_b$ is extremely small}, the
$\tfrac{1}{K_b^2}$-weighted statistical term dominates and \texttt{vanilla} is preferable; \textbf{when $K_b$ is larger}, the
geometric term dominates and \texttt{idct} becomes preferable.
If $\Delta_{\mathrm{st}}\le0$ or $\rho_C=0$, \cref{eq:specialize_compare} should instead be read directly as the
surrogate score difference, without a positive threshold.  We therefore use the threshold only as a design heuristic for
the two deployed coordinate candidates, not as a theorem over all rotations.

\paragraph{Random rotation.}
A Haar-random $\mathbf{R}$ typically densifies token correlations (large $\mathcal{L}_{\mathrm{st}}(\mathbf{R})$) and is
nearly orthogonal to $\mathbf{U}_C$ in high dimensions
($\langle \mathbf{R},\mathbf{U}_C\rangle_F\approx 0$), so it is dominated by $\mathbf{I}$ in the small-$K$ regime and by
$\mathbf{U}_C$ in the larger-$K$ regime.

\paragraph{Outcome.}
We implement a simple, reproducible proxy of \eqref{eq:specialize_threshold}: for very small backbones we keep
\texttt{vanilla} ($\mathbf{A}=\mathbf{I}$), and once the backbone reaches a modest size we switch to \texttt{idct}
($\mathbf{A}=\mathbf{U}_C$). In the main experiments, this uses \texttt{vanilla} at $K_b\in\{1,4,9\}$ and \texttt{idct}
at $K_b=16$, corresponding to total budgets $K\in\{4,9,16,25\}$.

\paragraph{Robustness of the threshold.}
The numeric cutoff \(K_b{=}16\) is not intended to be a universal constant.  What should transfer is the criterion in
\cref{eq:specialize_compare}, which compares the statistical-conditioning gap against the geometric-compatibility gain.
For a new dataset, encoder, or resolution, one can estimate \(\Delta_{\mathrm{st}}\) and \(\rho_C\) on a small unlabeled
calibration split and choose \texttt{vanilla} vs.\ \texttt{idct} by the sign of
\(\frac{1}{K_b^2}\Delta_{\mathrm{st}}-\beta\rho_C\).  The deployed choices in the main experiments are consistent with
this check: \texttt{vanilla} is better at \texttt{c3s7} (94.0 vs.\ 90.9), whereas \texttt{idct} is better at
\texttt{c4s9} (95.2 vs.\ 93.2).  \Cref{tab:app_coordinate_calibration} reports calibration results across three encoders
and two data distributions.

\section{Spatial residual domain}
\label{app:spatial_residual_domain}

\subsection{Residual-domain motivation}

Low-pass transform truncation is efficient for global structure but inevitably drops localized details.  To motivate the
spatial residual branch, first fix the DCT backbone induced by $(\mathbf{D}_N,\mathcal{S}_C)$ and define the corresponding
low-pass reconstruction in original token coordinates:
\begin{equation}
\begin{aligned}
\mathcal{P}(\mathbf{X})
&\triangleq
\mathbf{U}_{\mathbf{D}}^\top
\mathbf{P}_{\mathcal{S}_C}^\top
\mathbf{P}_{\mathcal{S}_C}
\mathbf{U}_{\mathbf{D}}\mathbf{X},\\
\mathbf{R}(\mathbf{X})
&\triangleq
\mathbf{X}-\mathcal{P}(\mathbf{X}),
\end{aligned}
\label{eq:specialize_residual_decomp}
\end{equation}
where $\mathbf{U}_{\mathbf{D}}=\mathbf{D}_N\otimes\mathbf{D}_N$, $\mathcal{P}(\mathbf{X})$ is the retained low-pass
component, and $\mathbf{R}(\mathbf{X})$ is the discarded residual.  Even when the low-pass subspace has high average
compressibility, task error can be dominated by localized components in this discarded residual.
To decide the residual domain, consider a single-channel residual $\mathbf{r}\in\mathbb{R}^L$ with at most $s$ nonzero
spatial entries.  For an orthonormal transform $\mathbf{U}\in\mathbb{R}^{L\times L}$, define its coherence with the
spatial basis as $\mu=\sqrt{L}\max_{i,j}|U_{ij}|$.
By Cauchy--Schwarz, each coefficient obeys
$|(\mathbf{U}\mathbf{r})_i|\le (\mu/\sqrt{L})\|\mathbf{r}\|_1
\le\mu\sqrt{s/L}\|\mathbf{r}\|_2$.
Squaring and summing over any $m$ coordinates gives
\begin{equation}
\max_{|\Omega|=m}
\|(\mathbf{U}\mathbf{r})_{\Omega}\|_2^2
\le
\frac{m\mu^2s}{L}\|\mathbf{r}\|_2^2,
\label{eq:specialize_incoherence}
\end{equation}
where $\Omega\subseteq[L]$ indexes selected transform coordinates.  For the orthonormal 2D DCT, $\mu\le2$.
When $m\mu^2s\ll L$, even the best $m$ transform coefficients retain only a small fraction of the residual energy.
This motivates Braco's spatial branch, which directly pools localized evidence to complement the low-pass backbone.

\subsection{Residual-pooling implementation}
\label{app:coder_details}

This subsection records the residual-pooling and projection details summarized in \cref{sec:coder}.  After the DCT backbone
and budget-dependent coordinate organization produce $\mathbf{Z}_b\in\mathbb{R}^{C^2\times D_v}$, a lightweight scorer
outputs residual logits $\boldsymbol{\ell}^{(s)}\in\mathbb{R}^{L}$ for each residual slot $s\in[S]$.  When enabled, a
normalized gradient-energy bias on the token grid is added to these logits before sparsemax; equivalently,
$\boldsymbol{\ell}^{(s)}$ denotes the biased logits in that case.  With sparsemax temperature $\tau>0$, the residual
weights are
\begin{equation}
\mathbf{w}^{(s)}
=
\mathrm{sparsemax}\!\left(\boldsymbol{\ell}^{(s)}/\tau\right),
\qquad
\mathbf{1}^{\top}\mathbf{w}^{(s)}=1,
\label{eq:coder_sparsemax}
\end{equation}
where $\mathrm{sparsemax}(\cdot)$ maps logits to the probability simplex with sparse support.  We anneal $\tau$ during
training to encourage sharper spatial selection.  The residual token is then the weighted spatial sum used in
\cref{eq:coder_pool}.

After concatenating backbone and residual tokens,
\begin{equation}
\mathbf{Z}=[\mathbf{Z}_b;\mathbf{Z}_r]\in\mathbb{R}^{(C^2+S)\times D_v},
\qquad
\mathbf{H}_{\mathrm{vis}}
=
p_{\mathrm{proj}}\!\left(\mathrm{Norm}(\mathbf{Z})\right),
\label{eq:coder_final}
\end{equation}
where $\mathrm{Norm}(\cdot)$ harmonizes token scales and $p_{\mathrm{proj}}$ maps visual-token features into the LLM
hidden size.

\section{Additional Experiments and Reproducibility}
\label{app:extended_experiments}

This appendix follows the same evidence chain as the main text: \cref{tab:app_design_validation_map} summarizes how each
diagnostic motivates a Braco design choice and where the corresponding end-to-end check appears.  We then clarify the
diagnostic setup and operating regime, give the exact Braco configurations, module-level costs, larger-input results,
training time, and hardware protocol.  Unless otherwise stated, ``Acc.'' denotes the same Vanilla-normalized aggregate used
in the main table.

\begin{table}[t]
  \centering
  \caption{\textbf{Design-validation map.}
  Each diagnostic identifies a failure mode, motivates a Braco design choice, and connects it to the corresponding
  end-to-end experiment.}
  \label{tab:app_design_validation_map}
  \footnotesize
  \setlength{\tabcolsep}{2.6pt}
  \renewcommand{\arraystretch}{1.08}
  \begin{tabular}{@{}>{\raggedright\arraybackslash}p{0.18\linewidth}
                  >{\raggedright\arraybackslash}p{0.25\linewidth}
                  >{\raggedright\arraybackslash}p{0.22\linewidth}
                  >{\raggedright\arraybackslash}p{0.25\linewidth}@{}}
    \toprule
    \textbf{Design axis} & \textbf{Failure mode diagnosed} & \textbf{Braco choice} & \textbf{End-to-end confirmation} \\
    \midrule
    Retained subspace &
    Tiny structured budgets may discard task-relevant visual directions. &
    Use a DCT low-frequency backbone. &
    DCT has higher energy/readability in \cref{fig:exp_compressibility_diag}; replacing it with spatial/Haar tokens drops
    Acc. in \cref{fig:exp_ablation_step}; functional rankings are validated in \cref{app:functional_validation}. \\
    Coordinate organization &
    Equal-information coordinates can differ in optimization and alignment. &
    Use a budget-dependent coordinate rule. &
    \Cref{fig:learnability_loss_curves_dev} isolates this effect within the same retained subspace; wrong coordinates or
    random rotations reduce Acc. in \cref{fig:exp_ablation_step}. \\
    Token identity &
    Transform coefficients no longer carry ordinary patch-position semantics. &
    Add basis-coordinate positional embeddings. &
    Replacing the basis-coordinate embedding with learned or 2D sine-cosine alternatives reduces Acc. in
    \cref{fig:exp_ablation_pe_alt}. \\
    Local evidence &
    A pure low-pass backbone can miss sparse localized evidence. &
    Allocate a small spatial residual budget. &
    The hybrid \texttt{c2s5} interface outperforms backbone-only and residual-only variants in
    \cref{fig:exp_ablation_cnsn,tab:app_residual_only_cost}. \\
    Compressor overhead &
    Learned resamplers can recover accuracy while adding module cost. &
    Keep the interface prompt-independent and lightweight. &
    Braco matches QueCC-level Acc. with lower compressor latency/FLOPs in
    \cref{tab:main_compressor_cost,tab:app_multi_budget_cost}. \\
    \bottomrule
  \end{tabular}
\end{table}

\subsection{Compression diagnostics and operating regime}
\label{app:compression_diagnostics}

\paragraph{Diagnostic interpretation.}
The compact diagnostic figure is placed in the main text as \cref{fig:exp_compressibility_diag}.  This appendix section
records how to read it and why the selected token budgets form the comparable extreme-compression regime.  The maps
visualize why a structured low-frequency DCT block is deployable: the retained coordinates cover the concentrated energy
region, whereas the same structured block in spatial or random coordinates misses most evidence.

\paragraph{Compressibility and readability protocol.}
The energy maps in \cref{fig:exp_compressibility_diag} and the energy curves in \cref{fig:energy_curves} are measured on
frozen visual-token grids.  After flattening the \(N\times N\) token grid along the token axis, the token-axis second moment
\(\mathbf{M}=\mathbb{E}[\mathbf{X}\mathbf{X}^{\top}]\) is used to compute the retained-energy ratio in
\cref{eq:unified_energy_retention}.  We compare spatial, DCT, Haar, and random orthonormal bases under structured
truncation, and include magnitude truncation as an adaptive top-\(K\) comparison that chooses the largest coefficients
after seeing each representation.  The KLT curve is reported as an oracle reference for the diagnostic distribution under
the same \(K\)-coordinate budget; Braco does not use this fitted basis.  It shows the headroom between fixed analytic bases
and a fitted orthonormal basis.

The CelebA~\cite{liu2015faceattributes} results in \cref{fig:exp_linear_probe_val,fig:exp_linear_probe_test} provide a controlled readability diagnostic,
separate from the learnability result in \cref{fig:learnability_loss_curves_dev}.  The dataset provides visually grounded binary
attributes, making it useful for testing whether frozen retained coordinates still support simple semantic readout at tiny
budgets.  For each basis and budget, we keep the visual encoder and compression map fixed and compare the retained-token
representations under the same linear-probe setup, reporting both validation and test curves.  The end-to-end VLM benchmarks
in \cref{tab:token-retain} remain the evidence for multimodal task performance.

\paragraph{Learnability protocol.}
\Cref{fig:learnability_loss_curves_dev} uses the pretraining objective, not CelebA.  We fix the low-frequency DCT backbone
subspace with \(K_b=C^2\), compare only subspace-preserving coordinate organizations (\texttt{vanilla}, \texttt{idct}, and
\texttt{randrot}), and split the stream into train/dev at 0.99/0.01.  We report time-to-threshold, defined as the first step
where held-out dev cross-entropy falls below 2.50, together with the final dev loss.  At \(K_b{=}4\), only \texttt{vanilla}
reaches the threshold
(1718 steps), while \texttt{idct} and \texttt{randrot} plateau above it.  At \(K_b{=}16\), \texttt{idct} reaches the same
threshold in 707 steps versus 860 for \texttt{vanilla} and 719 for \texttt{randrot}; at \(K_b{=}64\), \texttt{idct}
reaches it in 237 steps versus 590 for \texttt{vanilla} and 432 for \texttt{randrot}.

\paragraph{Extreme-budget regime.}
We treat compression ratios above \(23\times\) as the extreme-budget regime studied in this paper.  The main table therefore
uses \(K\in\{25,16,9,4\}\), corresponding to \(23\times\)--\(144\times\) compression from the 576-token LLaVA-v1.5~\cite{liu2024improved} visual
interface.  We do not report \(K{=}1\) or \(K{=}2\) because these settings are not cleanly comparable across key baselines:
QueCC is defined around square-grid outputs in its original setup and does not naturally support \(K{=}2\), while \(K{=}1\)
is degenerate for Braco because the method contains both a structured backbone and a residual branch.  We therefore set
\(K{=}4\) as the smallest comparable operating point and use \cref{tab:main_larger_visual_tokens} to study higher visual
resolutions and larger effective compression ratios.

\subsection{Main Braco configurations}
\label{app:braco_configs}

\paragraph{Configuration interpretation.}
\Cref{tab:app_braco_configs} gives the exact Braco instantiation behind the main-table budgets.  As the token budget
increases, Braco expands the low-frequency backbone and keeps a small residual budget for localized evidence; only the
25-token setting uses the \texttt{idct} coordinate organization, matching the budget-dependent behavior discussed in the
main text.

\subsection{Compression-module cost}
\label{app:module_cost}

We report module-level costs at the compression boundary used by each family of methods.  The main compressor comparison is
already reported in \cref{tab:main_compressor_cost}; this appendix adds stage-level and multi-budget breakdowns that are
not shown in the main text.

\paragraph{Stage-cost interpretation.}
\Cref{tab:app_stage_flops} decomposes Braco's compressor into its four implementation stages.  The transform and
basis-coordinate embedding stages are small and nearly identical across the representative allocations; the residual pooling
stage dominates the module cost, but remains at the GFLOP scale shown in the previous tables.

\paragraph{Residual-only interpretation.}
\Cref{tab:app_residual_only_cost} complements the main ablation in \cref{fig:exp_ablation_cnsn}.
The pairwise contrasts make the complementarity explicit: adding five residual tokens to the same \(C{=}2\) backbone
raises accuracy from 85.7 (\texttt{c2s0}) to 93.2 (\texttt{c2s5}), while adding the same backbone to five residual-only
tokens raises accuracy from 88.2 (\texttt{c0s5}) to 93.2 (\texttt{c2s5}) at essentially unchanged module FLOPs/latency.
Residual-only variants (\texttt{c0s5}, \texttt{c0s9}) are therefore weaker and not cheaper overall, supporting Braco's
joint parameterized backbone plus spatial residual design.

\begin{table}[t]
  \centering
  \caption{\textbf{Braco configurations used in the main table.}
  \texttt{c\{C\}s\{S\}} denotes a $C\times C$ low-frequency backbone and $S$ spatial residual tokens; PE denotes the
  basis-coordinate positional embedding.}
  \label{tab:app_braco_configs}
  \footnotesize
  \begin{tabular}{cccccc}
    \toprule
    Budget $K$ & Config. & Backbone $C^2$ & Residual $S$ & Coordinate org. & PE \\
    \midrule
    4 & \texttt{c1s3} & 1 & 3 & \texttt{vanilla} & yes \\
    9 & \texttt{c2s5} & 4 & 5 & \texttt{vanilla} & yes \\
    16 & \texttt{c3s7} & 9 & 7 & \texttt{vanilla} & yes \\
    25 & \texttt{c4s9} & 16 & 9 & \texttt{idct} & yes \\
    \bottomrule
  \end{tabular}
\end{table}

\begin{table}[h]
  \centering
  \caption{\textbf{Analytic FLOPs decomposition of Braco stages.}
  Values are reported in MFLOPs for the two representative allocations used in the main paper.}
  \label{tab:app_stage_flops}
  \footnotesize
  \begin{tabular}{lcccc}
    \toprule
    Allocation & Step 1 & Step 2 & Step 3 & Step 4 \\
    \midrule
    \texttt{c3s7} & 56.623 & 38.928 & 0 & 1236.293 \\
    \texttt{c4s9} & 56.623 & 38.928 & 0.262 & 1240.663 \\
    \bottomrule
  \end{tabular}
\end{table}

\begin{table}[h]
  \centering
  \caption{\textbf{Ablation cost: residual-only variants at the compressor boundary.}
  Residual-only variants are weaker and not cheaper overall than the hybrid backbone--residual allocation.}
  \label{tab:app_residual_only_cost}
  \footnotesize
  \begin{tabular}{lcccc}
    \toprule
    Method & Acc. (\%) & FLOPs (G) & Latency (ms) & Peak memory (MB) \\
    \midrule
    \texttt{c2s5} & 93.2 & 1.382 & 1.067 & 8.729 \\
    \texttt{c2s0} & 85.7 & 0.152 & 0.288 & 4.504 \\
    \texttt{c0s5} & 88.2 & 1.382 & 1.091 & 8.729 \\
    \texttt{c3s0} & 89.8 & 0.152 & 0.289 & 4.504 \\
    \texttt{c0s9} & 92.0 & 1.396 & 1.077 & 8.734 \\
    \bottomrule
  \end{tabular}
\end{table}

\begin{table}[h]
  \centering
  \caption{\textbf{Multi-budget Braco compressor cost.}
  Peak memory varies little across retained-token budgets.}
  \label{tab:app_multi_budget_cost}
  \footnotesize
  \begin{tabular}{lcccc}
    \toprule
    Method & \#Tokens & FLOPs (G) & Latency (ms) & Peak memory (MB) \\
    \midrule
    Braco & 4 & 1.375 & 1.060 & 8.727 \\
    Braco & 9 & 1.382 & 1.067 & 8.729 \\
    Braco & 25 & 1.396 & 1.197 & 8.734 \\
    \bottomrule
  \end{tabular}
\end{table}

\paragraph{Multi-budget cost interpretation.}
\Cref{tab:app_multi_budget_cost} shows that Braco's standalone compressor overhead changes only mildly from 4 to 25
retained tokens, excluding the 16-token row already reported in the main compressor-cost table.  This supports the
main-text cost analysis: the deployed interface remains lightweight across the extreme budget range, with no large overhead
increase at the larger operating points.

\subsection{Larger-input generalization and training time}
\label{app:larger_visual_tokens}
\label{app:larger_token_generalization}

We first report the complete larger-input accuracy and inference-cost comparison, then provide the corresponding wall-clock
training time for the same two-stage recipe.  These larger-input settings use the LLaVA-NeXT/AnyRes-style image pipeline~\cite{li2024llavaonevision}:
each high-resolution image is sliced into multiple visual views, which increases the number of vision-encoder tokens before
compression while keeping the downstream compression interface unchanged.

\begin{table}[t]
  \centering
  \caption{\textbf{Generalization to larger visual-token inputs.}
  FLOPs and latency include vision encoder, projector, and LLM prefill.}
  \label{tab:main_larger_visual_tokens}
  \footnotesize
  \setlength{\tabcolsep}{0.6pt}
  \renewcommand{\arraystretch}{1.04}
  \begin{tabular}{@{}llccrrrrrrrrrrr@{}}
    \toprule
    \textbf{Method} & \textbf{LLM} & \textbf{Input} & \textbf{Retain} & \textbf{GQA} & \textbf{MMB$^{\text{EN}}$} & \textbf{MMB$^{\text{CN}}$} & \textbf{MME} & \textbf{POPE} & \textbf{SQA} & \textbf{VQA$^{\text{Text}}$} & \textbf{MMVet} & \textbf{Acc.} & \textbf{FLOPs} & \textbf{Lat.} \\
    & & \textbf{tokens} & \textbf{tokens} & & & & & & & & & \textbf{(\%)} & \textbf{(T)} & \textbf{(ms)} \\
    \midrule
    Vanilla & Vicuna-7B & 576 & 576 & 62.9 & 65.5 & 60.7 & 1785 & 85.7 & 69.7 & 58.0 & 32.8 & 100.0 & 8.67 & 67.25 \\
    QueCC & Vicuna-7B & 576 & 16 (36$\times$) & 59.0 & 63.1 & 54.6 & 1668 & 83.5 & 70.6 & 52.8 & 28.8 & 93.9 & 1.36 & 63.22 \\
    Braco & Vicuna-7B & 576 & 16 (36$\times$) & 57.5 & 63.3 & 55.8 & 1704 & 82.4 & 69.5 & 51.9 & 29.8 & 94.0 & 1.25 & 40.59 \\
    \midrule
    Vanilla & Vicuna-7B & 2880 & 2880 & 63.7 & 66.2 & 59.7 & 1701 & 86.6 & 67.8 & 64.1 & 31.2 & 100.0 & 40.57 & 261.08 \\
    QueCC & Vicuna-7B & 2880 & 80 (36$\times$) & 62.0 & 65.1 & 57.5 & 1762 & 85.2 & 68.6 & 59.1 & 29.4 & 97.7 & 3.99 & 66.73 \\
    Braco & Vicuna-7B & 2880 & 80 (36$\times$) & 61.8 & 66.0 & 58.8 & 1801 & 83.9 & 69.1 & 57.3 & 29.8 & 98.1 & 3.45 & 44.36 \\
    \midrule
    Vanilla & Qwen2.5-3B & 729 & 729 & 63.5 & 74.7 & 73.5 & 1758 & 86.7 & 74.5 & 59.7 & 34.7 & 100.0 & 5.37 & 62.84 \\
    QueCC & Qwen2.5-3B & 729 & 16 (46$\times$) & 53.8 & 51.6 & 50.1 & 1484 & 83.4 & 68.8 & 42.5 & 17.8 & 77.2 & 0.97 & 84.77 \\
    Braco & Qwen2.5-3B & 729 & 16 (46$\times$) & 58.3 & 69.8 & 67.5 & 1723 & 83.6 & 75.6 & 53.3 & 28.7 & 93.1 & 0.96 & 49.56 \\
    \midrule
    Vanilla & Qwen2.5-3B & 1024 & 1024 & 61.7 & 74.7 & 72.5 & 1809 & 84.5 & 74.9 & 62.9 & 34.4 & 100.0 & 7.40 & 77.08 \\
    QueCC & Qwen2.5-3B & 1024 & 16 (64$\times$) & 51.2 & 50.3 & 49.0 & 1356 & 72.5 & 69.0 & 42.0 & 18.7 & 74.0 & 1.18 & 83.95 \\
    Braco & Qwen2.5-3B & 1024 & 16 (64$\times$) & 57.0 & 69.9 & 68.8 & 1744 & 81.6 & 74.2 & 54.0 & 28.2 & 92.6 & 1.10 & 52.04 \\
    \midrule
    Vanilla & Qwen2.5-3B & 3645 & 3645 & 63.8 & 75.0 & 72.7 & 1836 & 88.1 & 75.1 & 65.7 & 37.8 & 100.0 & 25.91 & 240.66 \\
    QueCC & Qwen2.5-3B & 3645 & 20 (182$\times$) & 50.0 & 41.1 & 39.4 & 1355 & 81.1 & 68.3 & 41.2 & 16.7 & 68.9 & 3.81 & 118.61 \\
    Braco & Qwen2.5-3B & 3645 & 20 (182$\times$) & 57.6 & 69.9 & 66.8 & 1742 & 84.3 & 75.5 & 53.8 & 29.5 & 90.8 & 3.50 & 88.64 \\
    \bottomrule
  \end{tabular}
\end{table}

\begin{table}[h]
  \centering
  \caption{\textbf{Training time on 8 NVIDIA A100 GPUs.}
  All entries use the same two-stage training recipe.}
  \label{tab:app_training_time}
  \footnotesize
  \begin{tabular}{lccc}
    \toprule
    Method & \#Vision tokens & Pre-training & Instruction-tuning \\
    \midrule
    Vanilla & 576 & 3.5h & 10h \\
    QueCC & 16 (36$\times$) & 0.7h & 7h \\
    Braco & 16 (36$\times$) & 0.4h & 6.5h \\
    Vanilla & 2880 & 14h & 28h \\
    QueCC & 80 (36$\times$) & 1.1h & 8.2h \\
    Braco & 80 (36$\times$) & 1h & 7.5h \\
    \bottomrule
  \end{tabular}
\end{table}

\paragraph{Larger-input interpretation.}
\Cref{tab:main_larger_visual_tokens} evaluates Braco along two axes: it remains effective when the
LLaVA-NeXT/AnyRes-style input grows beyond the 576-token LLaVA setting, and it also transfers from Vicuna-7B~\cite{vicuna2023} to the
smaller Qwen2.5-3B~\cite{yang2024qwen25technicalreport} backbone.  In the Vicuna-7B setting, increasing the input from 576 to 2880 visual tokens raises the uncompressed
full-pipeline cost from 8.67T FLOPs and 67.25ms to 40.57T FLOPs and 261.08ms.  At the same 36$\times$ compression ratio,
Braco improves normalized Acc. over QueCC (98.1 vs.\ 97.7), reduces FLOPs further (3.45T vs.\ 3.99T), and lowers
latency from 66.73ms to 44.36ms.

The Qwen2.5-3B rows show that QueCC's query-based compression is less favorable with a smaller LLM backbone: despite being
close to Braco in the 576-token Vicuna-7B main comparison, QueCC drops to 77.2/74.0 Acc. at 729/1024 input tokens and 68.9
Acc. at 3645 tokens.  A likely factor is that its prompt-conditioned compression queries provide a weaker semantic signal
under a smaller LLM, while the query/downsampling module adds enough fixed cost that latency exceeds the uncompressed
baseline at 729 and 1024 tokens (84.77/83.95ms vs.\ 62.84/77.08ms).  Braco avoids this mismatch, keeping 93.1/92.6 Acc.
at 46$\times$/64$\times$ compression while remaining faster than the corresponding uncompressed baselines.

The high-resolution, high-cost rows also clarify the practical significance of extreme compression.  The 576-token setting
mainly tests whether accuracy survives a very small retained-token budget; as the visual interface grows, the same small
budget removes a much larger uncompressed prefill burden and therefore yields larger end-to-end latency and FLOP savings.
Braco's 182$\times$ Qwen2.5-3B row still retains 90.8 Acc. while
cutting full-pipeline cost from 25.91T FLOPs and 240.66ms to 3.50T FLOPs and 88.64ms, indicating that the lightweight
backbone--residual interface remains useful when extreme compression is not just a benchmark stress test but a practical
deployment need.

\paragraph{Training-time interpretation.}
\Cref{tab:app_training_time} reports representative wall-clock training time under the shared two-stage recipe.  Braco is consistently no
slower than QueCC and is often faster, indicating that the lightweight interface does not introduce extra training burden
despite adding transform-domain parameterization and residual pooling.

\subsection{Hardware scaling and measurement protocol}
\label{app:hardware_protocol}
\label{app:hardware_scaling}

This subsection first states the shared measurement protocol used for the main LLaVA-v1.5-7B~\cite{liu2024improved} comparisons, then reports
additional A100/A800 batch-scaling measurements.

\paragraph{Metric and cost scope.}
For a benchmark set \(\mathcal{B}\), the main-table aggregate is the Vanilla-normalized accuracy
\begin{equation}
\label{eq:app_normalized_accuracy}
100|\mathcal{B}|^{-1}\sum_{b\in\mathcal{B}}s_b/s_b^{\mathrm{Vanilla}},
\end{equation}
where \(s_b\) is the method score on benchmark \(b\), and \(s_b^{\mathrm{Vanilla}}\) is the corresponding uncompressed
baseline score.  The FLOPs and latency columns in \cref{tab:token-retain} measure single-image full-pipeline prefill cost,
from the vision encoder through the projector and LLM prefill.

\begin{figure}[h]
  \centering
  \includegraphics[width=\columnwidth]{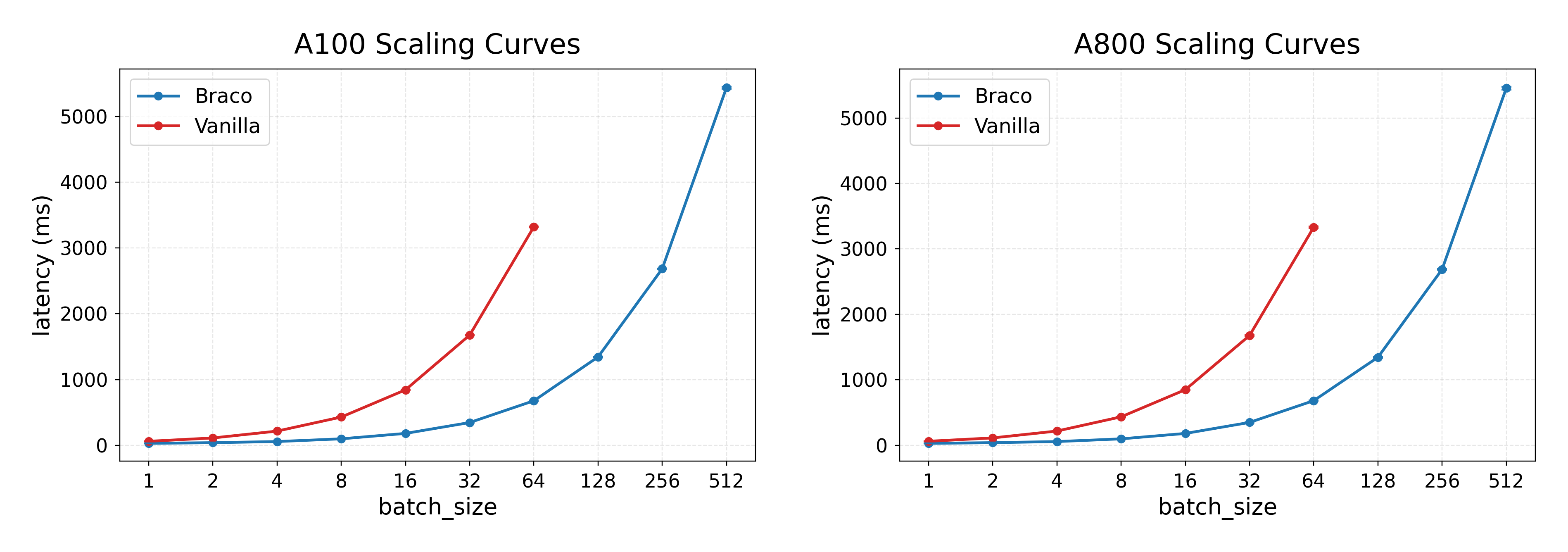}
  \caption{\textbf{Latency scaling curves on A100 and A800.}
  Comparison of Braco and Vanilla across batch sizes. Braco consistently achieves lower latency than Vanilla, and both
  methods exhibit increasing latency as batch size grows. Vanilla runs out of memory (OOM) at batch size 128 and above,
  so its curves stop at batch size 64. Curves show mean latency, with shaded bands indicating standard deviation over
  repeated measurements. This batch-scaling profile uses the same measurement protocol within each curve and is intended
  for scaling comparison.}
  \label{fig:a100_a800_scaling}
\end{figure}

\paragraph{Training and inference details.}
All main-table LLaVA-v1.5-7B~\cite{liu2024improved} numbers are produced by our own retraining and evaluation runs.
For every compressor baseline and Braco budget, we train a budget-specific checkpoint with the LLaVA two-stage
pipeline~\cite{liu2024improved}.
For both stages, the optimizer, learning-rate schedule, and weight decay follow the public LLaVA-v1.5-7B training recipe.
Both pre-training and instruction-tuning are run for one epoch.
In pre-training, the total batch size is 256 for 2181 steps, and the compressor/interface parameters and projector are
trained while the vision encoder and LLM remain frozen.
In instruction-tuning, the total batch size is 128 for 5198 steps, and the compressor/interface parameters, projector, and
LLM are trained.
The pre-training stage uses \texttt{blip\_laion\_cc\_sbu\_558k}; instruction-tuning uses \texttt{llava\_v1\_5\_mix665k};
both are the LLaVA-v1.5 data mixtures~\cite{liu2024improved}.  We then run the same benchmark and cost-measurement
scripts for each reproduced checkpoint.
Latency is measured on NVIDIA A100 GPUs, with additional A800 scaling results in \cref{fig:a100_a800_scaling} and
\cref{tab:latency_a100_a800}.
All inference costs are measured on a single GPU with a single image and a fixed-length prompt.
Main-table latency and FLOPs are averaged over 100 measured prefill runs after 20 warmup iterations; latency uses
CUDA events, and FLOPs use PyTorch-profiler estimates under the same input setting.
The batch-scaling measurements in \cref{fig:a100_a800_scaling} use the same prompt and implementation within each curve;
their absolute batch-1 latency is not intended to be directly compared with benchmark latency in \cref{tab:token-retain}.
We use CUDA 11.8, PyTorch 2.1.2, fp16 precision, 20 warmup iterations, and 100 measured runs.

\begin{table}[h]
  \centering
  \caption{\textbf{Latency comparison across batch sizes on A100 and A800.}
  Reported values are mean latency (ms) with standard deviation; OOM denotes Vanilla out-of-memory at batch size 128 and
  above.}
  \begin{subtable}[t]{0.48\textwidth}
    \centering
    \caption{\textbf{A100 latency comparison.}}
    \label{tab:a100_latency}
    \footnotesize
    \setlength{\tabcolsep}{3.5pt}
    \begin{tabular}{ccc}
      \toprule
      batch size & Braco & Vanilla \\
      \midrule
      1 & $30.71 \pm 0.39$ & $61.98 \pm 0.23$ \\
      2 & $40.94 \pm 0.12$ & $112.75 \pm 0.39$ \\
      4 & $58.07 \pm 0.52$ & $216.80 \pm 0.70$ \\
      8 & $99.40 \pm 0.43$ & $430.66 \pm 1.22$ \\
      16 & $181.64 \pm 0.54$ & $843.89 \pm 3.10$ \\
      32 & $346.87 \pm 0.55$ & $1675.45 \pm 3.43$ \\
      64 & $677.89 \pm 4.36$ & $3321.49 \pm 5.12$ \\
      128 & $1343.33 \pm 3.00$ & OOM \\
      256 & $2684.58 \pm 3.49$ & OOM \\
      512 & $5434.79 \pm 16.09$ & OOM \\
      \bottomrule
    \end{tabular}
  \end{subtable}
  \hfill
  \begin{subtable}[t]{0.48\textwidth}
    \centering
    \caption{\textbf{A800 latency comparison.}}
    \label{tab:a800_latency}
    \footnotesize
    \setlength{\tabcolsep}{3.5pt}
    \begin{tabular}{ccc}
      \toprule
      batch size & Braco & Vanilla \\
      \midrule
      1 & $30.25 \pm 0.08$ & $62.34 \pm 0.21$ \\
      2 & $41.04 \pm 0.04$ & $113.59 \pm 0.40$ \\
      4 & $57.84 \pm 0.07$ & $218.60 \pm 0.49$ \\
      8 & $99.06 \pm 0.17$ & $435.79 \pm 1.54$ \\
      16 & $181.82 \pm 0.57$ & $850.10 \pm 3.14$ \\
      32 & $350.15 \pm 1.11$ & $1677.62 \pm 5.63$ \\
      64 & $682.64 \pm 1.57$ & $3330.98 \pm 11.51$ \\
      128 & $1344.83 \pm 4.94$ & OOM \\
      256 & $2688.01 \pm 7.88$ & OOM \\
      512 & $5456.56 \pm 20.90$ & OOM \\
      \bottomrule
    \end{tabular}
  \end{subtable}
  \label{tab:latency_a100_a800}
\end{table}

\paragraph{Compute-resource accounting.}
\Cref{tab:app_training_time} reports measured wall-clock training time for representative Vanilla, QueCC, and Braco runs
on a cloud GPU platform with 8 NVIDIA A100 80 GB GPUs.
Multiplying by the worker count gives the run-level training budgets: 108, 61.6, and 55.2 A100 GPU-hours for the
576-token Vanilla, QueCC, and Braco runs, respectively, and 336, 74.4, and 68 A100 GPU-hours for the corresponding
2880-token runs.  We estimate total full-model training compute at approximately 4.5k A100 GPU-hours,
covering the main and larger-input runs, additional training seeds, and extended comparisons.

The benchmark and cost measurements use the same single-GPU inference workers: \cref{tab:token-retain,tab:main_larger_visual_tokens}
report per-image full-pipeline prefill FLOPs and latency, and \cref{tab:main_compressor_cost,tab:app_multi_budget_cost}
report compressor-boundary peak memory.  A800 80 GB resources are used only for additional batch-scaling latency in
\cref{fig:a100_a800_scaling,tab:latency_a100_a800}.  We recommend reserving approximately 2 TB of local storage for public
checkpoints, LLaVA data mixtures, reproduced checkpoints, derived features, and logs; no method-specific storage system is
required.  Preliminary debugging, calibration, and unused pilot variants used below 870 additional A100 GPU-hours
and are accounted for separately from the full-model training estimate.

\paragraph{Batch-scaling interpretation.}
\Cref{tab:latency_a100_a800} gives the values behind \cref{fig:a100_a800_scaling}, with A100/A800 splits in
\cref{tab:a100_latency,tab:a800_latency}; both GPUs show the same scaling trend.

\FloatBarrier
\subsection{Functional rankings and cross-encoder calibration}
\label{app:functional_validation}

\paragraph{Compressibility rankings.}
We evaluate four bases at each of six fixed budgets using a held-out two-layer MLP readout.  The coefficients
$\lambda,\beta,\gamma$ are held fixed, and the readability term $R$ uses only the probe-task training split.
No reported VLM validation/test labels are used.  All correlations are computed within a fixed budget.
The mean within-budget Spearman correlation with held-out MLP accuracy is $\rho=0.93$ for the combined compressibility
score $C$ in \cref{eq:unified_C}, compared with 0.67 for the energy term $E$ and 0.80 for the readability term $R$;
$C$ selects the best basis in all six budgets.  In the fully retrained 16-token basis ablation and the
9-token comparison, the basis ranking induced by $C$ agrees with the end-to-end ordering and selects the best
basis at both budgets.  The functional therefore guides basis selection across both nonlinear readout and full VLM training.

\paragraph{Learnability rankings.}
Across the coordinate candidates at $K_b\in\{4,16,64\}$, $\mathcal{L}_{\mathrm{learn}}$ in
\cref{eq:unified_learn_obj} has mean within-budget Spearman correlation $\rho=0.88$ with held-out dev-loss AUC.
At $K_b=4$, only coefficient coordinates reach the held-out dev-loss target; at $K_b=16$ and 64, iDCT reaches the
target 18\% and 60\% faster, respectively.  Separately, the coefficient-versus-iDCT choice induced by
\cref{eq:specialize_compare} matches the final-accuracy winner at $K_b\in\{4,9,16\}$
(\texttt{c2s5}/\texttt{c3s7}/\texttt{c4s9}), whereas the fixed random-rotation controls do not.
These results connect the learnability objective to both optimization speed and final accuracy, supporting
Braco's budget-dependent coordinate design.

\paragraph{Calibration protocol.}
We evaluate the coordinate criterion in \cref{eq:specialize_compare} on CLIP ViT-L/14-336,
SigLIP-SO400M/14-384~\cite{zhai2023siglip}, and SigLIP2-SO400M/16-512~\cite{tschannen2025siglip2}.
For each encoder and dataset, we sample 1,024 unlabeled images without replacement in three seed-defined resamples
(seeds 0/1/2), use the encoder's native square preprocessing, and scan $C=2,\ldots,16$ with $K_b=C^2$.
LLaVA-Pretrain (LLaVA-558K) determines the deployed threshold, while the DocVQA training split~\cite{mathew2021docvqa}
provides an OCR-heavy distribution-shift check.  No captions or downstream labels are used, and $\beta,\gamma$ are
fixed in the original CLIP setting and reused unchanged.  We define $K_b^\star$ as the first scanned budget for which
the criterion prefers iDCT.

\begin{table}[htbp]
  \centering
  \caption{\textbf{Coordinate-rule calibration across encoders and distributions.}
  Entries report the median crossover budget $K_b^\star$ and its range across three resamples of 1,024 unlabeled images.}
  \label{tab:app_coordinate_calibration}
  \footnotesize
  \begin{tabular}{lcc}
    \toprule
    Vision encoder / resolution & LLaVA-Pretrain $K_b^\star$ & DocVQA $K_b^\star$ \\
    \midrule
    CLIP ViT-L/14-336 & 16 (16--16) & 25 (16--25) \\
    SigLIP-SO400M/14-384 & 64 (64--64) & 81 (64--81) \\
    SigLIP2-SO400M/16-512 & 36 (36--36) & 49 (36--49) \\
    \bottomrule
  \end{tabular}
\end{table}

\Cref{tab:app_coordinate_calibration} shows that the calibration rule adapts to different encoder statistics while
remaining stable across resamples.  All three encoders yield identical thresholds across LLaVA-Pretrain resamples;
under the DocVQA shift, the threshold moves by at most one scanned square budget.  This gives Braco a practical,
label-free adaptation procedure: estimate the crossover from the visual-interface training distribution, or use a
small unlabeled sample from the target domain when the deployment distribution changes.

\subsection{Robustness to training seeds}
\label{app:training_robustness}

\paragraph{Repeated training.}
We evaluate Braco and QueCC at 16/9/4 tokens, and Braco and TokenPacker at 25/16/9 tokens, with three matched
training seeds per available method and budget.  Within each seed, the methods share the base checkpoint,
shuffled data order, and training schedule.  \Cref{tab:app_training_seeds} reports the sample mean and sample
standard deviation of normalized Acc. across the three runs.

\begin{table}[htbp]
  \centering
  \caption{\textbf{Accuracy across three matched training seeds.}
  Values are sample mean $\pm$ sample standard deviation.
  QueCC does not support 25 tokens on the fixed $24\times24$ lattice; TokenPacker is not included in the four-token
  repeated-training comparison.}
  \label{tab:app_training_seeds}
  \footnotesize
  \begin{tabular}{cccc}
    \toprule
    Tokens & Braco & QueCC & TokenPacker \\
    \midrule
    25 & $95.18\pm0.10$ & --- & $94.24\pm0.13$ \\
    16 & $94.12\pm0.10$ & $93.88\pm0.12$ & $93.70\pm0.15$ \\
    9 & $93.29\pm0.12$ & $93.02\pm0.14$ & $91.48\pm0.18$ \\
    4 & $91.28\pm0.15$ & $91.35\pm0.16$ & --- \\
    \bottomrule
  \end{tabular}
\end{table}

Braco preserves its accuracy advantage across three matched training runs: it leads QueCC at 16/9 tokens by
0.24/0.27 Acc. points and TokenPacker at 25/16/9 tokens by 0.94/0.42/1.81 points.
The Braco and QueCC standard deviations at 16/9 tokens are at most 0.14, showing low run-to-run variation.
At four tokens, the two methods remain effectively tied (a difference of $-0.07$), while Braco retains the
substantial latency advantage reported in \cref{tab:token-retain}.

\section{Limitations}
Braco is evaluated primarily on LLaVA-family vision-encoder$\rightarrow$LLM pipelines and standard single-image multimodal
benchmarks.  This scope reflects the need for fair, fully reproducible comparisons: Braco and the learned baselines are
trainable visual-interface modules, so matched comparisons require an open end-to-end training pipeline,
including the training code/recipe, data mixture, and checkpoints needed to retrain each method under the same conditions.
Its budget-dependent coordinate choice is calibrated for the tested encoders and resolutions, so new
backbones, video inputs, dense localization tasks, domain-specific distributions, irregular region features, or dynamic
visual tokens may require a small calibration sweep over the backbone--residual split, coordinate organization, or
structured basis.  The current study reports three-seed accuracy comparisons for the key Braco--QueCC and
Braco--TokenPacker settings (\cref{tab:app_training_seeds}), while the remaining accuracy comparisons use one trained
checkpoint per setting.

\section{Broader Impacts}
This paper studies extreme visual-token compression for vision--language models and proposes a lightweight token coder
that can reduce inference latency and memory footprint, enabling broader deployment on resource-constrained devices.
Improved efficiency may also lower the energy cost of multimodal systems, but it could facilitate more scalable
surveillance or automated content understanding when paired with high-capability models.  We do not introduce new data
sources or user-facing interaction mechanisms; the primary risks therefore mirror those of the underlying vision--language
models.  We encourage practitioners to follow existing responsible-deployment practices (e.g., data governance, access
control, and misuse monitoring) when applying our method in sensitive settings.

\section{Existing Assets and Licenses}
We use public third-party models, training data, benchmarks, and baseline implementations only under their original
licenses or terms of use, and this paper does not redistribute model weights, benchmark images, or dataset files.
The LLaVA codebase and LLaVA-v1.5 model family~\cite{liu2024improved} are released with Apache-2.0 code, while
the LLaVA-v1.5 and Vicuna-v1.5~\cite{vicuna2023} model weights follow the Llama 2 Community License.
The Qwen2.5-3B backbone~\cite{yang2024qwen25technicalreport} follows the Qwen Research License.
The LLaVA pretraining subset follows the LAION/CC/SBU image-source licenses, and the LLaVA instruction mixtures are
used under their posted dataset-card and source-dataset terms.
For evaluation assets, we use the original benchmark releases and cite the corresponding papers: MMBench data is
released under CC BY 4.0~\cite{liu2024mmbench}; ScienceQA code is MIT-licensed and its dataset is
CC BY-NC-SA 4.0~\cite{lu2022scienceqa}; MM-Vet code is Apache-2.0 and its dataset is CC BY-NC
4.0~\cite{yu2024mmvet}; TextVQA/VQA annotations are available under CC BY 4.0 where
applicable~\cite{singh2019textvqa}; GQA/Visual Genome assets are available under CC BY
4.0~\cite{hudson2019gqa}; and the POPE repository is MIT-licensed, with underlying MSCOCO-image terms where COCO
images are used~\cite{li2023pope}.  MME and any additional benchmark files are used under their posted release
terms~\cite{fu2023mme}.  Baseline methods are cited in the related work and experiments; any official code or
checkpoints used for reproduction are governed by their respective repository or model-card licenses, and methods
without reusable licensed code are reimplemented from the paper description.

\end{document}